\documentclass[letterpaper,10pt]{IEEEtran}

\usepackage[T1]{fontenc}
\usepackage[latin1]{inputenc}
\usepackage{float}
\usepackage{amsmath}
\usepackage[noadjust]{cite}
\usepackage{amssymb}
\usepackage{enumerate}
\usepackage{amsthm}
\usepackage{graphics,epsfig,psfrag}
\usepackage{subfigure,epsf,pstricks,subfloat}
\usepackage{bm}
\usepackage{url}
\usepackage{psfrag}
\usepackage{algorithm}
\usepackage{algorithmic}
\usepackage{verbatim}

\floatstyle{ruled}
\newfloat{algorithm}{tbp}{loa}

\algsetup{linenodelimiter=}

\theoremstyle{definition}

\title{Tensor Recovery from Noisy and Multi-Level Quantized Measurements}
\author{
\IEEEauthorblockN{Ren Wang, Meng Wang, Jinjun Xiong}
\thanks{R. Wang and M. Wang are with the Dept. of Electrical, Computer, and Systems Engineering,   Rensselaer Polytechnic Institute, Troy, NY.  Email:   \{wangr8, wangm7\}@rpi.edu. J. Xiong is with IBM Thomas J. Watson Research Center, Yorktown Heights, NY, Email: jinjun@us.ibm.com.}
}

\usepackage{verbatim}
\usepackage{graphicx}
\usepackage{epstopdf}
\newtheorem{theorem}{Theorem}
\newtheorem{lemma}{Lemma}

\newtheorem{defi}{Definition}

\begin{document}

\maketitle
\thispagestyle{empty} \pagestyle{empty}

\begin{abstract}
Higher-order tensors can represent scores in a rating system, frames in a video, and images of the same subject. In practice, the measurements are often highly quantized due to the sampling strategies or the quality of devices. Existing works on tensor recovery have focused on data losses and random noises. Only a few works consider tensor recovery from quantized measurements but are restricted to binary measurements.
This paper, for the first time,  addresses the problem of tensor recovery from multi-level quantized measurements. Leveraging the low-rank property of the tensor, this paper proposes a nonconvex optimization problem for tensor recovery. We provide a theoretical upper bound of the recovery error, which diminishes to zero when the sizes of dimensions increase to infinity. Our error bound significantly improves over the existing results in one-bit tensor recovery and quantized matrix recovery. A tensor-based alternating proximal gradient descent algorithm with a convergence guarantee is proposed to solve the nonconvex problem. Our recovery method can handle data losses and do not need the information of the quantization rule. The method is validated on synthetic data, image datasets, and music recommender datasets.
\end{abstract}


\section{Introduction}
Many practical datasets are  highly noisy and quantized, and recovering the actual values from quantized measurements finds applications in different domains. 
For example, users' preferences in rating systems are  represented by  a few scores (or even two scores in one bit  \cite{DPBW14}), which do not provide accurate characterization of preferences.
Due to  sensor issues or communication restrictions,  images and videos in some applications may have very low resolution \cite{WWX18}. Quantization is   applied to enhance the data privacy in power systems and sensor networks \cite{CMH16,GWWC18,REC13}. It is important to develop computationally efficient and reliable methods to recover the actual data from low-resolution measurements.

\cite{LTSM17}  estimates the data from one-bit measurements by linearizing the nonlinear quantizer. \cite{KNSE19} leverages the deep learning tool to recover the data. These approaches either require accurate parameter estimation or have high computational costs. \cite{PV2013,SL2015,ZYJ14}   recover data from a small number of quantized measurements, but the methods only apply to sparse signals. Low-rank matrices can characterize the intrinsic data correlations in user ratings, images, and videos \cite{BNS16,ZWL15}, and the low-rank property has been exploited  to recover the data from quantized measurements by solving a nonconvex constrained maximum likelihood estimation problem \cite{Bhaskar16,CZ13,DPBW14,GWWC18}.
For an $n\times n$ rank-$r$ matrix ($r \ll n$), the best achievable recovery error from quantized measurements is $O(\sqrt{\frac{r^3}{n}})$\footnote{We use the notations  $g= \mathcal{O}(n)$, $g=\Theta(n)$ if as $n$ goes to infinity, $g \leq c \cdot n$, $c_1 \cdot n \leq g \leq c_2 \cdot n$ eventually holds for some positive constants $c$, $c_1$ and $c_2$ respectively.}  \cite{Bhaskar16,GWWC18}. The recovery error diminishes to zero when the data size increases.

Practical datasets may contain additional correlations that cannot be captured by low-rank matrices. For instance, if an image or a frame of a video is vectorized, the spatial correlation is no longer preserved \cite{FGTL14}. In   recommendation systems, users ratings against objects vary under different contexts \cite{BLM11}, and a matrix presentation is not sufficient to characterize the structure. That motivates the usage of low-rank tensors, where a higher-order tensor contains data arrays with at least three dimensions. Tensors can represent  three-dimensional objects in generic object recognition \cite{SK03}, engagements on advertisements over time for behavior analysis \cite{BMR17}, gene expressions in the development process \cite{LZZZJ15}, etc. Moreover, tensor techniques are widely used in deep learning \cite{CS2016,MTO18}. 

Low-rank tensors with quantization noise exist in hyper-spectral data \cite{ATT18,LZLL19}, rating systems \cite{GPY19}, and the knowledge predicates \cite{ZZW16}. Existing works on low-rank tensor recovery mainly consider random noise or sparse noise \cite{CLKX13,RM2014,ZWZM13}, while only a few works \cite{ATT18,GPY19,LZLL19} consider tensor recovery from one-bit measurements. \cite{ATT18} unfolds the tensors to matrices and applies matrix recovery techniques. Ref. \cite{GPY19} recovers the tensor by solving a convex optimization problem and shows its recovery error is $O((\frac{r^{3K-3}K}{n^{K-1}})^{1/4})$, where $K$ is the number of tensor dimensions, and $n$ is the size per dimension. \cite{LZLL19} focuses on the case when a significant percentage of measurements are lost.

This paper for the first time studies low-rank tensor recovery from multi-level quantized measurements, while the existing work \cite{ATT18,GPY19,LZLL19} only consider one-bit measurements. We formulate the tensor recovery problem as a nonconvex optimization problem and proves that the recovery error with full observations under the known quantization rule is at most $O(\frac{r\sqrt{K\log(K)}}{\sqrt{n^{K-1}}})$, which decays to zero much faster than any existing results.
Moreover, we develop a computationally efficient algorithm to solve the nonconvex data recovery problem and prove that even with partial data losses,  our algorithm converges to a critical point from any initialization with at least sublinear convergence rate.
Lastly, all the existing work on recovery from quantized measurements assumes that the quantization rule is known to the recovery method except one low-rank matrix recovery work \cite{Bhaskar16}. We empirically extend   our method to recover the tensor from quantized measurements without directly knowing the quantization rule and demonstrate encouraging numerical results.

This paper is organized as follows. The problem formulation is introduced in Section \ref{gen_inst}. Section \ref{headings} discusses our approach and its recovery error. An efficient algorithm with the convergence guarantee is proposed in Section \ref{algorithm:tapgd}. Section \ref{experiment} records the numerical results. Section \ref{conclusion} concludes the paper. All the lemmas and proofs can be found in Appendix.

\subsection{Notation and preliminaries}
We use boldface capital letters to denote matrices (two-dimensional tensors), e.g., $\mathbf{A}$. Higher order tensors (three or higher dimensions) are denoted by capital calligraphic letters, e.g., $\mathcal{X}$. $\mathcal{X} \in \mathbb{R}^{n_1 \times n_2 \times \dots \times n_K}$ represents a $K$-dimensional tensor with the size of the $i$-th dimension equaling to $n_i, i \in [K]$, where $[K] = \{1,2,\dots,K\}$. $\mathcal{X}_{i_1,i_2,\dots,i_K}$ denotes the $(i_1,i_2,\dots,i_K)$-th entry of $\mathcal{X}$. $\mathbf{X}_{(k)} \in \mathbb{R}^{n_k \times (n_1 \dots n_{k-1} n_{k+1} \dots n_K)}$ is the mode-$k$ matricization of $\mathcal{X}$, which is formed by unfolding $\mathcal{X}$ along its $k$-th dimension. 

Let $a_i \in \mathbb{R}^{n_i}, \forall i \in [K]$ be $K$ vectors. Then $\mathcal{A} = a_1 \circ \dots \circ a_K$ is a $K$-dimensional tensor with $\mathcal{A}_{i_1,i_2,\dots,i_K} = {a_1}_{i_1} {a_2}_{i_2} \dots {a_K}_{i_K}$. Here $\circ$ is called the outer product. The rank of $\mathcal{X}$ \cite{FL18} is defined as 
\begin{equation}\label{rank}
\begin{aligned}
&\text{rank}(\mathcal{X}) = \min\{R: \mathcal{X} = \\&\sum_{i=1}^{R} \mathbf{A_1}_i \circ \mathbf{A_2}_i \circ \dots \circ \mathbf{A_K}_i, \mathbf{A_k} \in \mathbb{R}^{n_k \times R}, k \in [K]\},
\end{aligned}
\end{equation}
where $\mathbf{A_k}_i$ is the $i$-th column of $\mathbf{A_k}$. $\mathbf{A_1} \circ \mathbf{A_2} \circ \dots \circ \mathbf{A_K}$ is equivalent to $\sum_{i=1}^{R} \mathbf{A_1}_i \circ \mathbf{A_2}_i \circ \dots \circ \mathbf{A_K}_i$. We use $\mathbf{A_k} \odot \mathbf{A_p}$ to represent the Khatri-Rao product \cite{KB09} of $\mathbf{A_k} \in \mathbb{R}^{n_k \times r}, \mathbf{A_p} \in \mathbb{R}^{n_p \times r}$. We have $\mathbf{A_k} \odot \mathbf{A_p} = [\mathbf{A_k}_1 \bigotimes \mathbf{A_p}_1, \mathbf{A_k}_2 \bigotimes \mathbf{A_p}_2, \dots, \mathbf{A_k}_r \bigotimes \mathbf{A_p}_r]$, where $\mathbf{A_k}_i \bigotimes \mathbf{A_p}_i \- = \-[(\mathbf{A_k}_{i})_1\mathbf{A_p}_i^T,\-(\mathbf{A_k}_{i})_2\mathbf{A_p}_i^T,\- \dots,\-(\mathbf{A_k}_{i})_{n_k}\mathbf{A_p}_i^T]^T\- \in \-\mathbb{R}^{n_kn_p \times 1},\-\forall i \in [r]$.

The Frobenius norm of the tensor $\mathcal{X}$ is defined as $\|\mathcal{X}\|_F = \sqrt{\sum_{i_1 = 1}^{n_1}\sum_{i_2 = 1}^{n_2}\dots\sum_{i_K = 1}^{n_K}\mathcal{X}_{i_1,i_2,\dots,i_K}^2}$.

\section{Problem formulation}
\label{gen_inst}
Let $\mathcal{X}^* \in \mathbb{R}^{n_1 \times n_2 \times \dots \times n_K}$ denote the actual data that are represented by a $K$-dimensional tensor. Let $\|\cdot\|_\infty$ denote the entry-wise infinity norm. We assume that the maximum value of $\mathcal{X}^*$ is bounded by a positive constant $\alpha$, i.e., $\|\mathcal{X}^*\|_\infty \le \alpha$. We further assume $\text{rank}(\mathcal{X}^*)\le r$.

\begin{figure}
	\centering
	\includegraphics[width=1\linewidth]{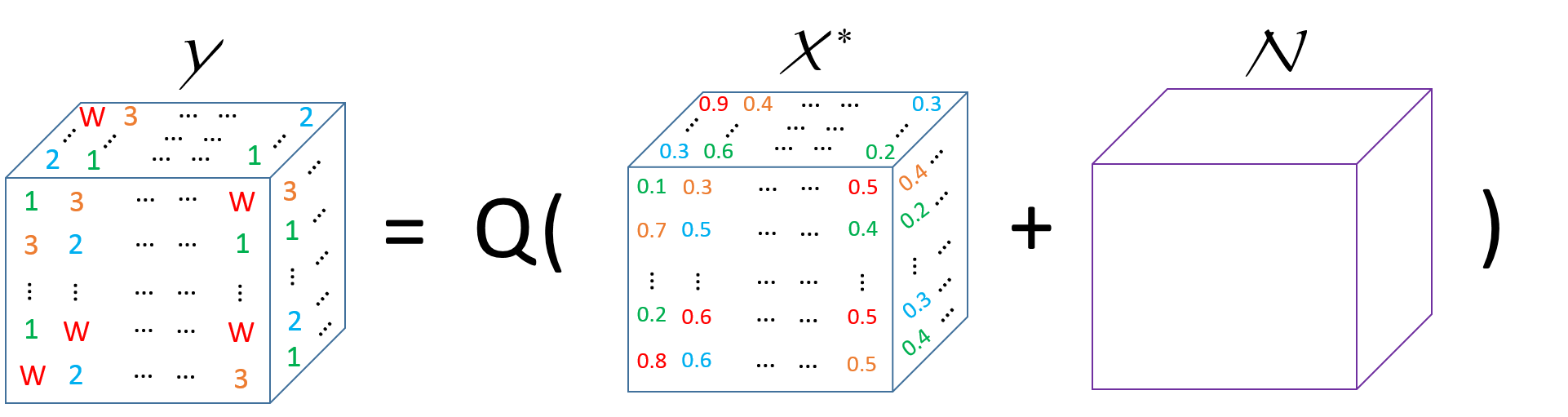}
	\caption{Quantization model ($K=3$).}
	\label{fig1}
\end{figure}

Each entry of $\mathcal{X}^*$ is mapped to one of a few possible values with certain probabilities through the quantization process. To model this probabilistic mapping, let $\mathcal{N} \in \mathbb{R}^{n_1 \times n_2 \times \dots \times n_K}$ denote a noise tensor with i.i.d. entries drawn from a known cumulative distribution function $\Phi(x)$. Given the quantization boundaries $\omega_0^* < \omega_1^*< \dots < \omega_W^*$, the noisy data $\mathcal{X}_{i_1,i_2,\dots,i_K}^* + \mathcal{N}_{i_1,i_2,\dots,i_K}$ ($i_j \in [n_j], j \in [K]$) can be quantized to $W$ values based on the following rule,
\begin{equation}\label{quant}
\begin{aligned}
&\mathcal{Y}_{i_1,i_2,\dots,i_K} = Q(\mathcal{X}_{i_1,i_2,\dots,i_K}^* + \mathcal{N}_{i_1,i_2,\dots,i_K}) = l \ \ \\& \textrm{if} \ \omega_{l-1}^*<\mathcal{X}_{i_1,i_2,\dots,i_K}^* + \mathcal{N}_{i_1,i_2,\dots,i_K}\le\omega_{l}^*, \ l \in [W],
\end{aligned}
\end{equation}
where $Q$ is an operator that maps a real value to one of $W$ values. We choose $\omega_0^* = -\infty$ and $\omega_W^* = \infty$. $\mathcal{Y}_{i_1,i_2,\dots,i_K}$ is the $(i_1,i_2,\dots,i_K)$-th entry of the quantized measurements $\mathcal{Y} \in [W]^{n_1 \times n_2 \times \dots \times n_K}$. When $W=2$, $\mathcal{Y}$ reduces to the one-bit case \cite{GPY19}. In general, $\mathcal{Y}$ is a $\log_2 W$-bit tensor. The quantization process of a three-dimensional tensor is visualized in Fig.~\ref{fig1}. 

The probability that $\mathcal{Y}_{i_1,i_2,\dots,i_K} = l$ given $\mathcal{X}_{i_1,i_2,\dots,i_K}^*, \omega_{l-1}^*, \omega_l^*$ is expressed by $f_l(\mathcal{X}_{i_1,i_2,\dots,i_K}^*, \omega_{l-1}^*, \omega_l^*)$, where
\begin{equation}\label{f_l}
\begin{aligned}
&f_l(\mathcal{X}_{i_1,i_2,\dots,i_K}^*, \omega_{l-1}^*, \omega_l^*)\\&=P(\mathcal{Y}_{i_1,i_2,\dots,i_K} = l|\mathcal{X}_{i_1,i_2,\dots,i_K}^*, \omega_{l-1}^*, \omega_l^*) \\&= \Phi(\omega_l^*-\mathcal{X}_{i_1,i_2,\dots,i_K}^*)-\Phi(\omega_{l-1}^*-\mathcal{X}_{i_1,i_2,\dots,i_K}^*),
\end{aligned}
\end{equation}
and $\sum_{l=1}^Wf_l(\mathcal{X}_{i_1,i_2,\dots,i_K}^*, \omega_{l-1}^*, \omega_l^*)= \Phi(\infty-\mathcal{X}_{i_1,i_2,\dots,i_K}^*)-\Phi(-\infty-\mathcal{X}_{i_1,i_2,\dots,i_K}^*) = 1$.
The probability description (\ref{f_l}) follows from the same formula as those in \cite{Bhaskar16,GWWC18}, except that the entries are from a higher order tensor. We assume $\Phi(x)$ is monotonously increasing. The monotonously increasing property holds for the cumulative distribution functions of many distributions. Examples include: (1) Probit model with $\Phi(x)=\Phi_{\rm norm}(x/\sigma)$, where $\Phi_{\rm norm}$ is the cumulative distribution function of a standard Gaussian distribution; (2) Logistic model with $\Phi(x)=\Phi_{\rm log}(x/\sigma)=\frac{1}{1+e^{-x/\sigma}}$.

We also consider the general setup that there exists missing data in the measurements, i.e., only measurements with indices belonging to the observation set $\Omega$ are available, while all the other measurements are lost.
The question we will address in this paper is as follows. Given the partial observations $\mathcal{Y}_\Omega$ and the noise distribution $\Phi$, how can we estimate the original tensor $\mathcal{X}^*$?

We remark that this problem formulation can be applied in different domains. In the user voting systems, data can be represented as $\{\text{users} \times \text{scoring objects} \times \text{contexts}\}$ \cite{BLM11}, which is a three-dimensional tensor. The scores from the reviewers are highly quantized \cite{DPBW14}. By solving the quantized tensor recovery problem, one can obtain the actual preferences of the reviewers. In video processing, the measurements can be represented as $\{\text{rows of a frame} \times \text{columns of a frame} \times \text{different frames}\}$. The measurements can be highly quantized due to the sensing process and the objective is to recover the data \cite{BLL2010,ZJWB09}. A similar idea also applies to low-quality image recovery \cite{FGTL14,WWX18}. Images from the same subject can be represented by $\{\text{rows of an image} \times \text{columns of an image} \times \text{different images}\}$.

\section{Tensor recovery from quantized measurements}
\label{headings}

We propose to estimate tensor $\mathcal{X}^*$, boundaries $\omega_{1}^*, \omega_{2}^*, \cdots, \omega_{W-1}^*$
using a constrained maximum likelihood approach. The negative log-likelihood function is given by
\begin{equation}\label{eqn:llh}
\begin{aligned}
&F_\Omega(\mathcal{X},\omega_1, \omega_2,\cdots, \omega_{W-1}) = -\frac{n_1n_2\cdots n_K}{|\Omega|}\sum_{(i_1,i_2,\cdots,i_K) \in \Omega}\\&\sum_{l=1}^W \boldsymbol{1}_{[\mathcal{Y}_{i_1,i_2,...,i_K}=l]}\log(f_l(\mathcal{X}_{i_1,i_2,...,i_K},\omega_{l-1}, \omega_l)),
\end{aligned}
\end{equation}
where $\boldsymbol{1}_{[B]}$ is an indicator function that takes value `$1$' if the event $B$ is true and value `$0$' otherwise. $|\Omega|$ denotes the cardinality of $\Omega$. (\ref{eqn:llh}) is a convex function when $f_l$ is a log-concave function. When $\omega_l^*$'s are unknown, we estimate $\mathcal{X}^*$, $\omega_l^*$'s by $\hat{\mathcal{X}}$, $\hat{\omega}_l$'s, where
\begin{equation}\label{problem_gen}
\begin{aligned}
&(\hat{\mathcal{X}},\hat{\omega}_1,\hat{\omega}_2,\cdots,\hat{\omega}_{W-1})\\& = {\arg\min}_{\mathcal{X} , \omega_l, \forall l \in [W-1]} F_\Omega(\mathcal{X},\omega_1, \omega_2,\cdots, \omega_{W-1}) \: \:\: \: \\&\textrm{s.t. }   \mathcal{X},\omega_1, \omega_2,\cdots, \omega_{W-1} \in \mathcal{S}_{f\omega},
\end{aligned}
\end{equation}
where
\begin{equation}\label{eqn:c1}
\begin{aligned}
\mathcal{S}_{f\omega}:=&\{\mathcal{X} \in \mathbb{R}^{n_1\times n_2 \times ... n_K}, \omega_l, \forall l \in [W-1]:  \\& \|\mathcal{X}\|_{\infty}\le\alpha, \text{rank}(\mathcal{X}) \le r,\\& \omega_0 < \omega_1 < \omega_2 < \cdots < \omega_W-1 < \omega_W\}.
\end{aligned}
\end{equation}

Most existing work on quantized data recovery consider the special case that the quantization boundaries are known \cite{DPBW14,GWWC18,GPY19} only except for \cite{Bhaskar16}. In this case, 

(\ref{problem_gen}) can be simplified to
\begin{equation}\label{problem1}
\hat{\mathcal{X}} = {\arg\min}_{\mathcal{X}} F_\Omega(\mathcal{X}, \omega_1^*,\omega_2^*,...,\omega^*_{W-1}) \: \:\: \: \textrm{s.t. }   \mathcal{X} \in \mathcal{S}_f,
\end{equation}
where
\begin{equation}\label{eqn:c2}
\mathcal{S}_f:=\{\mathcal{X} \in \mathbb{R}^{n_1\times n_2 \times ... n_K}: \|\mathcal{X}\|_{\infty}\le\alpha, \text{rank}(\mathcal{X}) \le r\},
\end{equation}

We remark that both \eqref{problem_gen}-\eqref{eqn:c1} and (\ref{problem1})-\eqref{eqn:c2} are nonconvex problems since $\mathcal{S}_{f\omega}$ and $\mathcal{S}_f$ are nonconvex sets.

\cite{GPY19} studies the case with known bin boundaries in (\ref{problem1})-\eqref{eqn:c2}. It focuses on the special case that $W=2$ and relaxes the low-rank constraint in $\mathcal{S}_f$ with a convex M-norm constraint. \cite{Bhaskar16,GWWC18} consider minimizing a negative log-likelihood function subject to a low-rank constraint, which is similar to (\ref{problem1}), but are restricted to quantized matrix recovery.

Similar to \cite{Bhaskar16,DPBW14}, we first define two constants $\gamma_{\alpha}$ and $L_{\alpha}$ for analysis in the case boundaries are all known constants. For simplicity, we denote $f_l(x, \omega_{l-1}^*,\omega_l^*)$ by $f_l(x)$.
\begin{equation}
\begin{aligned}
&\gamma_{\alpha} = \min_{l\in[W]}\inf_{|x|\le2\alpha}\{\frac{\dot{f}_l^2(x)}{f_l^2(x)}-\frac{\ddot{f}_l(x)}{f_l(x)}\},\\& L_{\alpha} = \max_{l\in[W]}\sup_{|x|\le2\alpha}\{\frac{|\dot{f}_l(x)|}{f_l(x)}\},
\end{aligned}
\end{equation}
where $\dot{f}_l$ and $\ddot{f}_l$ are the first and second order derivatives of $f_l$. Note that $\ddot{f}_l - \dot{f}_lf_l \geq 0$ if $f_l$ is log-concave, and $\ddot{f}_l - \dot{f}_lf_l > 0$ if $f_l$ is strictly log-concave. One can check that if $\Phi$ is monotonously increasing, then $f_l$ is strictly log-concave. Thus, $\gamma_{\alpha} > 0$ in our setup. We also remark that $L_\alpha$ and $\gamma_\alpha$ are bounded by some fixed constants when both $\alpha$ and $f_l$ are given. Taking the logistic model as an example \cite{GWWC18,Bhaskar16}, we have
\begin{equation}\label{log_para}
\begin{aligned}
&\gamma_{\alpha} =  \min_{l\in[W]}\inf_{|x|\le2\alpha}\frac{1}{\sigma^2}[\Phi_{\text{log}}(\frac{\omega_l-x}{\sigma})(1-\Phi_{\text{log}}(\frac{\omega_l-x}{\sigma}))\\&+\Phi_{\text{log}}(\frac{\omega_{l-1}-x}{\sigma})(1-\Phi_{\text{log}}(\frac{\omega_{l-1}-x}{\sigma}))] \\& L_{\alpha} = \\& 1/[2\sigma\min_{l\in[W]}\inf_{|x|\le2\alpha} \{ \Phi_{\text{log}}(\frac{\omega_l-x}{\sigma})-\Phi_{\text{log}}(\frac{\omega_{l-1}-x}{\sigma})\}]
\end{aligned}
\end{equation}
where  $L_\alpha$ and $\gamma_\alpha$ depend on $\sigma$  and $W$.
It is also easy to check that $\gamma_\alpha, L_\alpha >0$ from (\ref{log_para}).

We next state our main result that characterizes the recovery error   when there are no data losses and the quantization boundaries are known, i.e., the accuracy of the solution to \eqref{problem1} when $\Omega$ is the full observation set.

\begin{theorem}\label{theorem1}
Suppose $\omega_l^*$'s are given, and $\Omega$ contains all the indices. 
	 $\mathcal{X}^* \in \mathcal{S}_f$, and $f_l(x)$ is strictly log-concave in $x$, $\forall l \in [W]$.   Then, with probability at least $1-\delta$, $\delta \in [0,1]$, any global minimizer $\hat{\mathcal{X}}$ of (\ref{problem1}) satisfies
	\begin{equation}\label{mainresult_jl}
	\|\hat{\mathcal{X}}-\mathcal{X}^*\|_F/\sqrt{n_1n_2...n_K} \le \min (2\alpha,U_{\alpha})
	\end{equation}
	where
	\begin{equation}\label{U_jl}
	U_{\alpha} = \frac{4r}{\gamma_\alpha}\sqrt{\frac{8L_\alpha^2((\sum_{k=1}^{K}n_k)\log(4K/3)+\log(2/\delta))}{n_1n_2...n_K}},
	\end{equation}
\end{theorem}

Theorem \ref{theorem1} establishes the upper bound of the recovery error when the measurements are noisy and quantized. $L_\alpha, \delta, \gamma_\alpha$ are all constants. Specifically, when $n_1,n_2,\dots, n_K$ are all in the order of $n$,
\begin{equation}\label{thm_order}
\|\hat{\mathcal{X}}-\mathcal{X}^*\|_F/\sqrt{n_1n_2...n_K} \le O(\frac{r\sqrt{K\log(K)}}{\sqrt{n^{K-1}}}),
\end{equation}
The right-hand-side of (\ref{thm_order}) diminishes to zero when $n$ increases to infinity. Note that the Frobenius norm of $X^*$ is in the same order of $\sqrt{n_1n_2...n_K}$. By dividing the actual error by  $\sqrt{n_1n_2...n_K}$, we have that the left-hand side of (\ref{thm_order}) is in the same order of the   relative error   $\|\hat{\mathcal{X}}-\mathcal{X}^*\|_F/\|\mathcal{X}^*\|$, which is a commonly used normalized error measure. Therefore, the relative recovery error is sufficiently close to zero when the size  of the tensor is large enough. 

Note that the recovery error depends on $W$ implicitly because $W$ affects $L_\alpha$ and $\gamma_\alpha$. 
It might seem counter-intuitive that the recovery error is not a monotone function of $W$. That is because we consider all the possible selections of bin boundaries for a given $W$ when computing $L_\alpha$ and $\gamma_\alpha$. A larger $W$ does not necessarily lead to more information in the quantized measurements. For example, if two bin boundaries are  very close to each other, almost no data would be mapped to this bin, and the effective number of quantization levels is less than $W$  (think of the extreme case when $\omega_1 = \omega_{W-1}$). This is why $W$ does not appear directly in the recovery bound. Of course in practice, in most cases, larger $W$ (more bits) will provide us more information about the real data, and thus increase the performance.

\paragraph{Recovery enhancement over the existing work on one-bit tensor recovery.}
Tensor recovery from one-bit measurements has been studied in \cite{GPY19}. \cite{GPY19} relaxes the nonconvex low-rank constraint with a convex M-norm constraint, and the resulting recovery method has an error bound of $O((\frac{r^{3K-3}K}{n^{K-1}})^{1/4})$. 
In contrast, our recovery error bound decays to zero faster than the approach in \cite{GPY19} for any $K\ge 2$. For example, the recovery error bound in (\ref{thm_order}) is $O(\frac{r}{n})$ when $K=3$, while the bound is $O((\frac{r^{3/2}}{n^{1/2}}))$ in \cite{GPY19}. 

\paragraph{Recovery enhancement over quantized matrix recovery.}
Tensor $\mathcal{X}$ can be unfolded to its mode-$k$ matricization $\mathbf{X}_{(k)}$ along the $k$-th dimension. When the size of each dimension is $\Theta(n)$, the sizes of the two dimensions of $\mathbf{X}_{(k)}$ are $\Theta(n)$ and $\Theta(n^{K-1})$, respectively. Thus, we can compare the recovery error in (\ref{thm_order}) with the results obtained by applying quantized matrix recovery methods on $\mathbf{X}_{(k)}$. \cite{DPBW14,GWWC18,Bhaskar16} provide the theoretical analyses of matrix recovery from quantized measurements. The recovery error are in the order of $O(\sqrt{\frac{\bar{r}^3}{\bar{N}}})$ and $O((\frac{\bar{r}}{\bar{N}})^{1/4})$ in \cite{Bhaskar16} and \cite{DPBW14}, respectively, where $\bar{r}$ is the rank of the matrix, and $\bar{N}$ is the smallest size in the two dimensions. Here in $\mathbf{X}_{(k)}$, $\bar{r}$ is smaller or equal to $r$, and $\bar{N}$ is $\Theta(n)$.  The best existing bound of  quantized matrix recovery is $O(\sqrt{\frac{\bar{r}}{\bar{N}}})$ \cite{GWWC18}, which means an error bound of $O(\sqrt{\frac{\bar{r}}{n}})$ if we unfold the tensor to $\mathbf{X}_{(k)}$. Note that the error order in (\ref{thm_order}) has a power of $K-1$ in its denominator. For example, the recovery error is $O(\frac{r}{n})$ by (\ref{thm_order}) when $K = 3$. Since $\bar{r},r \ll n$, the recovery error of (\ref{thm_order}) decays to zero faster than $O(\sqrt{\frac{\bar{r}}{n}})$ by \cite{GWWC18} for the matricization case when $K \ge 3$.

\paragraph{Reduction to the matrix case.}
When reduced to the matrix case, i.e., $K=2$, (\ref{thm_order}) shows that the quantized matrix recovery has an error bound of $O(\frac{r}{\sqrt{n}})$, which is close to the smallest error bound $O(\sqrt{\frac{r}{n}})$ \cite{GWWC18}. The difference $\sqrt{r}$ comes from a technical nuclear norm relaxation in the proof and can be ignored when $r$ is a constant.

\section{Alternating proximal gradient descent based on tensors}\label{algorithm:tapgd}
In this section, we develop a fast algorithm named Tensor-based Alternating Proximal Gradient Descent (TAPGD) to solve the nonconvex problem (\ref{problem_gen}) with the convergence guarantee.

Since $\text{rank}(\mathcal{X})\le r$, there exists $\mathbf{A_k}\in \mathbb{R}^{n_k \times r}, \forall k \in [K]$, such that $\mathcal{X} = \mathbf{A_1} \circ \mathbf{A_2} \circ \dots \circ \mathbf{A_K}$. Then we change the rank constraint into a penalty function $\frac{\lambda}{2}\|\mathcal{X} - \mathbf{A_1} \circ \mathbf{A_2} \circ \dots \circ \mathbf{A_K}\|_F^2$ in the objective, where $\lambda$ is a positive constant. The equality constraint holds when $\lambda$ goes to infinity. Note that $\mathcal{X} = \mathbf{A_1} \circ \mathbf{A_2} \circ \dots \circ \mathbf{A_K}$ is in the form of CANDECOMP/PARAFAC (CP) decomposition \cite{Harshman70}. Unlike matrix decomposition and the other major tensor decomposition method (Tucker decomposition \cite{Tucker66}), CP decomposition has a very weak requirement for the uniqueness of tensor factors. A sufficient condition for CP decomposition to be unique is that the summation of independent column numbers in $\mathbf{A_k}, k=1,2,\cdots,K$ is larger or equal to $2r+K-1$, which often holds true. In contrast, Tucker decomposition is generally not unique, and is usually computationally expensive to update its core tensor.

We revise $\mathcal{S}_{f\omega}$ to add constraints that quantization boundaries shall not be too close to avoid trivial solutions in practice. The resulting  feasible set is 
\begin{equation}\label{boundaries}
\begin{aligned}
&\mathcal{S}_\omega = \{ \mathcal{X}, \omega_{1}, \omega_{2},\cdots,\omega_{W-1}: \alpha_{\text{low}} \le \omega_1 \le \omega_2 - \kappa_2,\\&  \omega_{l-1} + \kappa_l \le \omega_l \le \omega_{l+1} - \kappa_{l+1},~ \forall l \in \{2,,3,...,W-2\},  \\& \omega_{W-2} + \kappa_{W-1} \le \omega_{W-1} \le \alpha_{\text{upper}},\|\mathcal{X}\|_{\infty}\le\alpha \}, 
\end{aligned}
\end{equation}
where $\kappa_l, \forall l \in \{2, 3,\cdots,W-1\}$ are some positive numbers that can be chosen using hyperparameter tuning or simply set as small positive constants, and $\kappa_1=\kappa_W=0$. $\alpha_{\text{low}}, \alpha_{\text{upper}}$ are two constants that provide the lower and upper bound of the boundaries, which could be chosen as $-\alpha$ and $\alpha$, or estimates computed in different applications. 
The revised problem of (\ref{problem_gen}) is shown as follows
\begin{equation}\label{problem2}
\begin{aligned}
&(\hat{\mathcal{X}},\hat{\omega}_1,\hat{\omega}_2,\cdots,\hat{\omega}_{W-1}) =\\& {\arg\min}_{\mathcal{X}, \mathbf{A_k}, k \in [K], \omega_l, l\in [W-1]} F_\Omega(\mathcal{X},\omega_1,\omega_2,\cdots,\omega_{W-1}) \\&+ \frac{\lambda}{2}\|\mathcal{X} - \mathbf{A_1} \circ \mathbf{A_2} \circ \dots \circ \mathbf{A_K}\|_F^2 + \\&\Psi_1(\mathcal{X})+\sum_{l=1}^{W-1}\Psi_2(\omega_l) 
\end{aligned}
\end{equation}
where 
\begin{equation}\label{Pro1}
\begin{aligned}
&\Psi_1(\mathcal{X})= \left \{ \begin{array}{rcl}
\infty & \mbox{if}~ \|\mathcal{X}\|_{\infty} > \alpha \\& \\ 0 & \text{otherwise}
\end{array} \right.
\\&\Psi_2(\omega_l)= \left \{ \begin{array}{rcl}
\infty & \mbox{if}~ \omega_l > \min(\omega_{l+1} - \kappa_{l+1}, \alpha_{\text{upper}}) \\& \text{or}~~ \omega_l < \max(\omega_{l-1} + \kappa_l, \alpha_{\text{low}})  \\ 0 & \text{otherwise}
\end{array} \right. 
\end{aligned}
\end{equation}
 $\Psi_1(\mathcal{X})$ is transformed by the constraint $\|\mathcal{X}\|_{\infty}\le\alpha$. $\Psi_2(\omega_l)$ is transformed by the constraints on $\omega_l$ in $\mathcal{S}_\omega$. 
Let
\begin{equation}\label{fun_H}
\begin{aligned}
H = &F_\Omega(\mathcal{X},\omega_1, \omega_2,\cdots, \omega_{W-1}) \\&+ \frac{\lambda}{2}\|\mathcal{X} - \mathbf{A_1} \circ \mathbf{A_2} \circ \dots \circ \mathbf{A_K}\|_F^2.
\end{aligned}
\end{equation}
Then we solve (\ref{problem2}) using proximal gradient method \cite{BST14}. The main steps of the proximal gradient method include updating $\mathcal{X}, \mathbf{A_k}, k \in [K], \omega_l, l \in [W-1]$ by using the gradient descent method on $H$, and projecting the result to $\mathcal{S}_{\omega}$. Since for $\forall k \in [K]$
\begin{equation}\label{equal}
\begin{aligned}
&\|\mathcal{X} - \mathbf{A_1} \circ \mathbf{A_2} \circ \dots \circ \mathbf{A_K}\|_F =\\& \|\mathbf{X}_{(k)} - \mathbf{A_k}(\mathbf{A_K}\odot \dots \odot \mathbf{A_{k+1}}\odot \mathbf{A_{k-1}}\odot \dots \mathbf{A_1})^T\|_F,
\end{aligned}
\end{equation}
the partial gradients of $H$ with respect to $\mathbf{A_k}$ and $\mathcal{X}$ can be calculated by
\begin{equation}\label{gradient1}
\begin{aligned}
\nabla_{\mathbf{A_k}} H = (\mathbf{A_k}(\mathbf{B_k})^T - \mathbf{X}_{(k)}) \mathbf{B_k}, \forall k \in [K],
\end{aligned}
\end{equation}
\begin{equation}\label{gradient2}
\begin{aligned}
\nabla_{\mathcal{X}} H =& \nabla_\mathcal{X} F_\Omega(\mathcal{X},\omega_1, \omega_2,\cdots, \omega_{W-1}) \\&+ \lambda (\mathcal{X} - \mathbf{A_1} \circ \mathbf{A_2} \circ...\circ \mathbf{A_K}), 
\end{aligned}
\end{equation}
where $\mathbf{B_k} = \mathbf{A_K}\odot ...\odot \mathbf{A_{k+1}} \odot \mathbf{A_{k-1}} \odot ...\odot \mathbf{A_1}$.  For any $(i_1,i_2,\cdots,i_K) \in \Omega$,
\begin{equation}
\begin{aligned}
&\nabla_\mathcal{X} F_\Omega(\mathcal{X},\omega_1, \omega_2,\cdots, \omega_{W-1})_{i_1,i_2,\dots,i_K} \\&= \frac{\dot{\Phi}(\omega_l-\mathcal{X}_{i_1,i_2,\dots,i_K})-\dot{\Phi}(\omega_{l-1}-\mathcal{X}_{i_1,i_2,\dots,i_K})}{\Phi(\omega_l-\mathcal{X}_{i_1,i_2,\dots,i_K})-\Phi(\omega_{l-1}-\mathcal{X}_{i_1,i_2,\dots,i_K})}.
\end{aligned}
\end{equation}
Otherwise, for any $(i_1,i_2,\cdots,i_K) \notin \Omega$
\begin{equation}
\begin{aligned}
\nabla_\mathcal{X} F_\Omega(\mathcal{X},\omega_1, \omega_2,\cdots, \omega_{W-1})_{i_1,i_2,\dots,i_K} = 0.
\end{aligned}
\end{equation}
The partial derivative of $H$ with respect to $\omega_l$ is shown as follows
\begin{equation}\label{gradient3}
\begin{aligned}
&\nabla_{\mathbf{\omega_{l}}} H = (\sum_{(i_1,i_2,\cdots,i_K) \in \Omega}\\& \frac{\boldsymbol{1}_{[\mathcal{Y}_{i_1,i_2,...,i_K}=l+1]}\dot{\Phi}(\omega_l-\mathcal{X}_{i_1,i_2,\dots,i_K})}{\Phi(\omega_{l+1}-\mathcal{X}_{i_1,i_2,\dots,i_K})-\Phi(\omega_{l}-\mathcal{X}_{i_1,i_2,\dots,i_K})} )\\&-( \sum_{(i_1,i_2,\cdots,i_K) \in \Omega}\\&\frac{\boldsymbol{1}_{[\mathcal{Y}_{i_1,i_2,...,i_K}=l]}\dot{\Phi}(\omega_l-\mathcal{X}_{i_1,i_2,\dots,i_K})}{\Phi(\omega_l-\mathcal{X}_{i_1,i_2,\dots,i_K})-\Phi(\omega_{l-1}-\mathcal{X}_{i_1,i_2,\dots,i_K})}), 
\end{aligned}
\end{equation}
The step sizes of the gradient descent are selected as
\begin{equation}\label{stepsize}
\begin{aligned}
&\tau_{\mathbf{A_k}} = \frac{1}{\|(\mathbf{B_k})^T\mathbf{B_k}\|},\forall k \in [K],\\& \tau_{\mathcal{X}} = \frac{1}{\frac{1}{\sigma^2\beta^2}+\lambda},\\&\tau_{\omega_l} = \frac{\sigma^2\beta^2}{\sqrt{G_{l}}+\sqrt{G_{l+1}}},\forall l \in [W-1],
\end{aligned}
\end{equation}
where $\|(\mathbf{B_k})^T\mathbf{B_k}\|$, $\frac{1}{\sigma\beta}+\lambda$, $\frac{\sqrt{G_{l}}+\sqrt{G_{l+1}}}{\sigma^2\beta^2}$ are Lipschitz constants of $\nabla_{\mathbf{A_k}} H$, $\nabla_{\mathcal{X}} H$, and $\nabla_{\omega_l}H$. $G_{l},G_{l+1}$ are the number of entries in $\mathcal{Y}_\Omega$ that equal to $l$ and $l+1$, respectively. Here $\beta$ is a small positive value that satisfies $\Phi(\omega_l-\mathcal{X}_{i_1,i_2,\dots,i_K}) \geq \Phi(\omega_{l-1}-\mathcal{X}_{i_1,i_2,\dots,i_K}) + \beta$. This holds true since $\mathcal{X}_{i_1,i_2,\dots,i_K}, \omega_l, \omega_{l-1}$ are all bounded, $\omega_l$ is larger than $\omega_{l-1}$, and $\Phi$ is a monotonously increasing function.
After updating $\mathcal{X}$, the algorithm sets $\mathcal{X}_{i_1,i_2,...,i_K}$ to $\alpha$ if $\mathcal{X}_{i_1,i_2,...,i_K} > \alpha$, and sets $\mathcal{X}_{i_1,i_2,...,i_K}$ to $-\alpha$ if $\mathcal{X}_{i_1,i_2,...,i_K} < -\alpha$. 
After updating $\omega_l$, the algorithm sets $\omega_l = \min(\omega_{l+1} - \kappa_{l+1}, \alpha_{\text{upper}})$ if $\omega_l > \min(\omega_{l+1} - \kappa_{l+1}, \alpha_{\text{upper}})$, and sets $\omega_l = \max(\omega_{l-1} + \kappa_l, \alpha_{\text{low}})$ if $\omega_l < \max(\omega_{l-1} + \kappa_l, \alpha_{\text{low}})$.

The algorithm is initialized by first estimating $\omega_l^*$'s according to the applications or simply setting $\omega_l^0 = \frac{2\alpha l}{W}-\alpha$ if no information is available, and then setting
\begin{equation}
\mathcal{X}_{i_1,i_2,...,i_K}^0 = \left \{ \begin{array}{rcl} \frac{\omega_l^0 + \omega_{l-1}^0}{2}, & \mbox{if}~ 1<\mathcal{Y}_{i_1,i_2,...,i_K}=l<W. \\ \frac{\alpha + \omega_{W-1}^0}{2}, & \mbox{if}~ \mathcal{Y}_{i_1,i_2,...,i_K}=W. \\ \frac{-\alpha + \omega_{1}}{2}, & \mbox{if}~ \mathcal{Y}_{i_1,i_2,...,i_K}=1. \\ 0, & (i_1,i_2,\cdots,i_K) \not\in \Omega.
\end{array} \right.
\end{equation}
$\mathbf{A_k}^0 \in \mathbb{R}^{n_k \times r},\forall k \in [K]$ are obtained through the decomposition of $\mathcal{X}^0$. The details of TAPGD is shown in Algorithm \ref{algorithm}. Note that when the quantization boundaries $\omega_l^*$'s are known, TAPGD can be revised easily by removing steps 14 - 20 from Algorithm \ref{algorithm}.
\begin{algorithm}[h]
	\caption{Tensor Based Alternating Proximal Gradient Descent (TAPGD)}
	\begin{algorithmic}[1]\label{algorithm}
		\REQUIRE Quantized tensor $\mathcal{Y}_\Omega \in \mathbb{R}^{n_1 \times n_2 \times \dots \times n_K}$, 
		initialization $\omega_l^0, l \in [W-1],$ tensor $\mathcal{X}^{0}$, matrices $\mathbf{A_k}^0, k \in [K]$, parameters $r$, $\sigma$, $\beta$, $\kappa_l, l \in [W],$ $\alpha_{\text{upper}},$ $\alpha_{\text{low}}$.
		\FOR{$t = 0, 1, 2, \dots, T$}
		\FOR{$k = 1, 2, \dots, K$}
		\STATE{$\mathbf{B_k}^{t-1} = \mathbf{A_K}^{t-1}\odot ...\odot \mathbf{A_{k+1}}^{t-1} \odot \mathbf{A_{k-1}}^{t} \odot ...\odot \mathbf{A_1}^{t}$.}
		\STATE{$\nabla_{\mathbf{A_k}} H = (\mathbf{A_k}^{t-1}(\mathbf{B_k}^{t-1})^T - \mathcal{X}_{(k)}) \mathbf{B_k}^{t-1}$, and $\tau_{\mathbf{A_k}}^{t-1} = 1/\|(\mathbf{B_k}^{t-1})^T\mathbf{B_k}^{t-1}\|$.}
		\STATE{$\mathbf{A_k}^{t} = \mathbf{A_k}^{t-1} - \tau_{\mathbf{A_k}}^{t-1}\nabla_{\mathbf{A_k}} H$.}
		\ENDFOR
		\STATE{$\nabla_{\mathcal{X}} H = \nabla_{\mathcal{X}} F_\Omega(\mathcal{X}^{t-1},\omega_1^{t-1},\omega_2^{t-1},\cdots,\omega_{W-1}^{t-1}) + \lambda (\mathcal{X}^{t-1} - \mathbf{A_1}^{t} \circ \mathbf{A_2}^{t} \circ...\circ \mathbf{A_K}^{t})$, \\and $\tau_{\mathcal{X}}^{t-1} = \frac{1}{\frac{1}{\sigma\beta}+\lambda}$.}
		\STATE{$\mathcal{X}^{t} = \mathcal{X}^{t-1} - \tau_{\mathcal{X}}^{t-1}\nabla_{\mathcal{X}} H$.}
		\FOR{$i_j = 0, 1, 2, \dots, n_j, \forall j \in [K]$}
		\STATE{\textbf{if} $\mathcal{X}_{i_1,i_2,...,i_K}^{t} > \alpha$, \textbf{then} set $\mathcal{X}_{i_1,i_2,...,i_K}^{t}=\alpha$.}
		\STATE{\textbf{else if} $\mathcal{X}_{i_1,i_2,...,i_K}^{t} < -\alpha$, \textbf{then} set $\mathcal{X}_{i_1,i_2,...,i_K}^{t}=-\alpha$.}
		\STATE{\textbf{end if}}
		\ENDFOR
		\FOR{$l = 1, 2, \dots, W-1$}
		\STATE{Calculate $\nabla_{\mathbf{\omega_{l}}} H$ according to (\ref{gradient3}), and $\tau_{\omega_l}^{t-1}=\frac{\sigma^2\beta^2}{\sqrt{G_{l}}+\sqrt{G_{l+1}}}$.}
		\STATE{$\omega_l^{t} = \omega_l^{t-1} - \tau_{\omega_l}^{t-1}\nabla_{\mathbf{\omega_{l}}} H$.}
		\STATE{\textbf{if} $\omega_l^t > \min(\omega_{l+1}^{t-1} - \kappa_l, \alpha_{\text{upper}})$, \textbf{then} set $\omega_l^t = \min(\omega_{l+1}^{t-1} - \kappa_l, \alpha_{\text{upper}})$.}
		\STATE{\textbf{else if} $\omega_l^t < \max(\omega_{l-1}^t + \kappa_l, \alpha_{\text{low}})$, \textbf{then} set $\omega_l^t = \max(\omega_{l-1}^t + \kappa_l, \alpha_{\text{low}})$.}
		\STATE{\textbf{end if}}
		\ENDFOR
		\ENDFOR
		\RETURN $\mathcal{X}, \omega_1, \omega_2, \cdots, \omega_{W-1}$.
	\end{algorithmic}
\end{algorithm}

To improve the recovery performance, one can multiple $\lambda$ by a small constant larger than one in each iteration. This provides a better numerical result than fixing $\lambda$ in all iterations.
The complexity of TAPGD in each iteration is $O(Krn_1n_2\dots n_K)$. The convergence of TAPGD is summarized in Theorem \ref{theorem2}.
\begin{theorem}\label{theorem2}
	Assume that the sequence $\{\mathbf{A_k}^{t}\}$ generated by Algorithm \ref{algorithm} is bounded. Then TAPGD globally converges to a critical point of (\ref{problem2}) from any initial point, and the convergence rate is at least $O(t^{\frac{\theta - 1}{2\theta - 1}})$, for some $\theta \in (\frac{1}{2},1)$. 
\end{theorem}
Theorem \ref{theorem2} indicates a sublinear convergence of TAPGD. One way to satisfy the requirement of bounded sequence is to scale the factorized variables so that $\|\mathbf{A_1}\|_F=\|\mathbf{A_2}\|_F=\cdots=\|\mathbf{A_K}\|_F$ after each iteration. We find TAPGD performs well numerically without the additional steps.

\section{Experiments}\label{experiment}
We conduct simulations on synthetic data, image data, and data from an in-car music recommender system \cite{BLM11} in this section. The recovery performance is measured by $\|\mathcal{X}^*-\tilde{\mathcal{X}}\|_F^2/\|\mathcal{X}^*\|_F^2$, where $\tilde{\mathcal{X}}$ is our estimation of $\mathcal{X}^*$. $K=3$ in both tests on synthetic data and real data. We set $T=200$. All the results are averaged over $100$ runs. The simulations are run in MATLAB on a 3.4GHz Intel Core i7 computer.
\subsection{Synthetic data}
A rank-$r$, three-dimensional tensor is generated as follows. We first generate $\mathbf{A_1} \in \mathbb{R}^{n_1 \times r}$ with entries sampled independently from a uniform distribution in $[-0.5,0.5]$, $\mathbf{A_2} \in \mathbb{R}^{n_2 \times r}$, and $\mathbf{A_3} \in \mathbb{R}^{n_3 \times r}$ with each entry sampled independently from a uniform distribution in $[0,1]$. Then we obtain the tensor by calculating $\mathbf{A_1} \circ \mathbf{A_2} \circ \mathbf{A_3}$ and scaling all the values to $[-1,1]$. The entries of $\mathcal{N}$ are i.i.d. generated from the Gaussian distribution with mean $0$ and the standard deviation of $0.25$. We choose $W = 2$ (one-bit) and $4$ (2-bit) in our experiments. When $W = 2$, $\omega_0^* = -\infty$, $\omega_1^* = 0$, $\omega_2^* = \infty$. When $W = 4$, $\omega_0^* = -\infty$, $\omega_1^* = -0.4$, $\omega_2^* = 0$, $\omega_3^* = 0.4$, $\omega_4^* = \infty$.

\begin{figure}[h]
	\centering
	\includegraphics[trim=7 7 20 20,clip,width=1\linewidth]{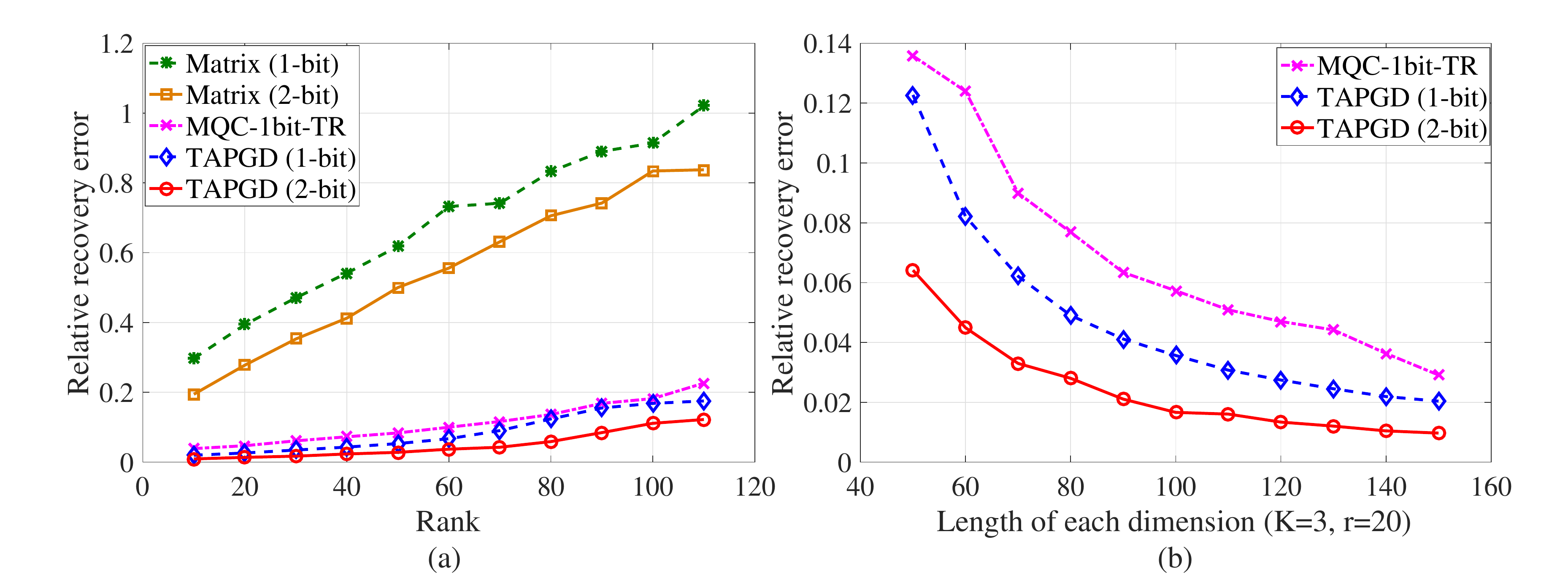}
	\caption{(a) Relative recovery error when rank changes (b) Relative recovery error when dimension changes}
	\label{fig2}
\end{figure}

\begin{figure}[h]
	\centering
	\includegraphics[trim=5 5 10 10,clip,width=0.75\linewidth]{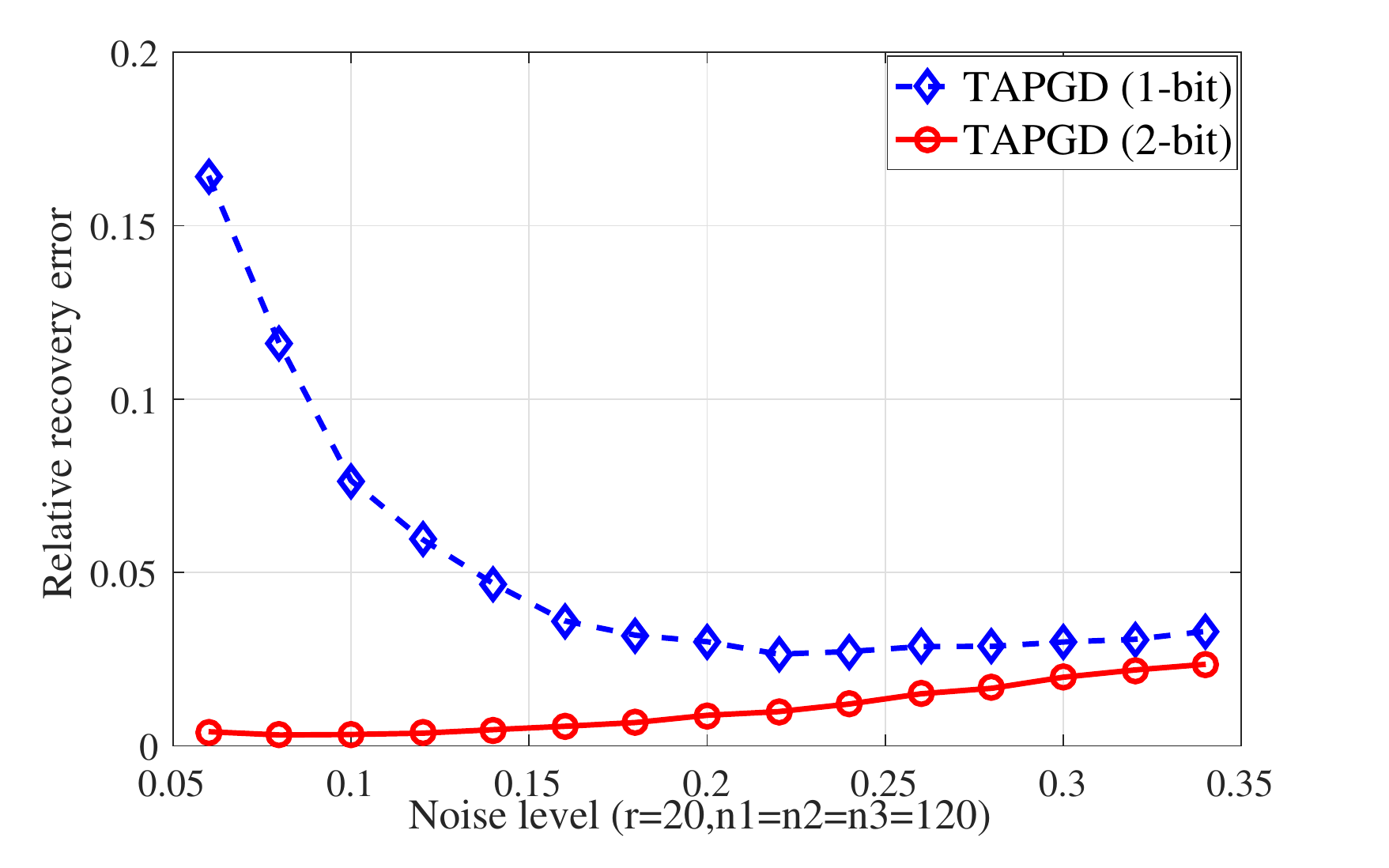}
	\caption{Relative recovery error when noise level changes}
	\label{fig3}
\end{figure}

Fig.~\ref{fig2} compares TAPGD with M-norm constrained one-bit tensor recovery (MNC-1bit-TR) method \cite{GPY19} and the quantized matrix recovery method \cite{GWWC18}. We remark that MNC-1bit-TR can only deal with one-bit measurements and requires solving a convex optimization problem. We vary one of the rank, dimension, noise level while fixing other parameters. $n1=n2=n3=120$ when we only vary rank and noise level. Fig.~\ref{fig2} demonstrates that the relative recovery error increases when rank increases or dimension decreases. The results are consistent with the theoretical analysis in Section \ref{headings}. The results also show that TAPGD has the best performance among all these methods. Moreover, performance improves when the number of bits increases. When the noise level increases, the relative recovery error first decreases, and then increases. The reason behind this is that the noise is considered as part of the quantization process, and plays the role of adding measurement uncertainty. The problem without noise (measurement uncertainty) is ill-posed in the sense that there may be an infinite number of solutions.  Large error under low noise corresponds to the case that the observations are nearly noise-free (especially for lower bits). The same behavior exists in the 2-bit curve when the noise level is smaller than $0.08$. Large error under high noise comes from the mask of the high randomness. From Fig.~\ref{fig3}, a low noise level does not necessarily mean low recovery error, and vice versa.

\subsection{Image data}
\begin{figure}[h]
	\centering
	\includegraphics[trim=7 0 20 20,clip,width=1\linewidth]{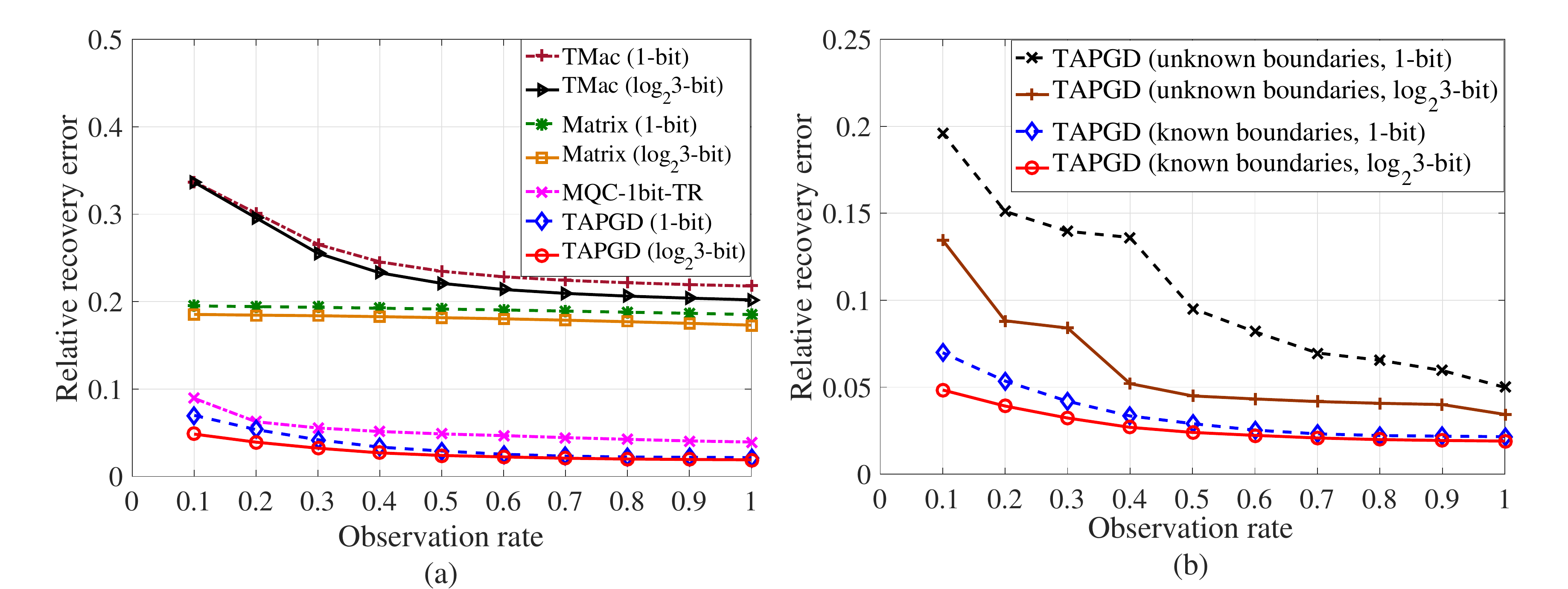}
	\caption{(a) Relative recovery error when the observation rate changes (b) Relative recovery error of unknown boundaries}
	\label{fig4}
\end{figure}

We test our method on the Extend Yale Face Dataset B \cite{SBK01,LHK05}. The dataset contains $192\times 168$ pixel face images from $38$ different people. Each person has $64$ images with different poses and various illumination. We pick two objects to form a $192\times 168 \times 128$ three-dimensional tensor. All entries are scaled to $[0,1]$. We add $\mathcal{N}$ with i.i.d. entries generated from the Gaussian distribution with mean $0$ and the standard deviation of $0.3$. When $W = 2$, $\omega_0^* = -\infty$, $\omega_1^* = 0.4$, $\omega_2^* = \infty$. When $W = 3$, $\omega_0^* = -\infty$, $\omega_1^* = 0.2$, $\omega_2^* = 0.4$, $\omega_3^* = \infty$. Fig.~\ref{fig4} (a) compares TAPGD with MNC-1bit-TR, the quantized matrix recovery method, and a nonconvex low-rank tensor recovery method named tensor completion by parallel matrix factorization (TMac) \cite{XYH13}. Note that MNC-1bit-TR models the quantization process like our approach, while TMac does not model quantization and treats the data as general noisy measurements. It shows that the relative recovery error decreases when the percentage of the observation increases, and TAPGD obtains the best performance among all the methods. Fig.~\ref{fig4} (b) compares the recovery error when the bin boundaries are known and unknown to the recovery algorithm. When the boundaries are unknown, the initial point is uniformly chosen from $[0.1, 0.6]$ for $\omega_1$ when $W=2$, and $[0.1,0.3],[0.2,0.6]$ for $\omega_{1},\omega_{2}$, respectively when $W=3$. $\alpha_{\text{upper}}, \alpha_{\text{low}}$ are selected as $0.6, 0.1$. $\kappa_l$ is set to $0.1$ for $\forall l \in [W-1]$.

\begin{figure}[h]
	\centering
	\includegraphics[trim=5 5 10 10,clip,width=0.75\linewidth]{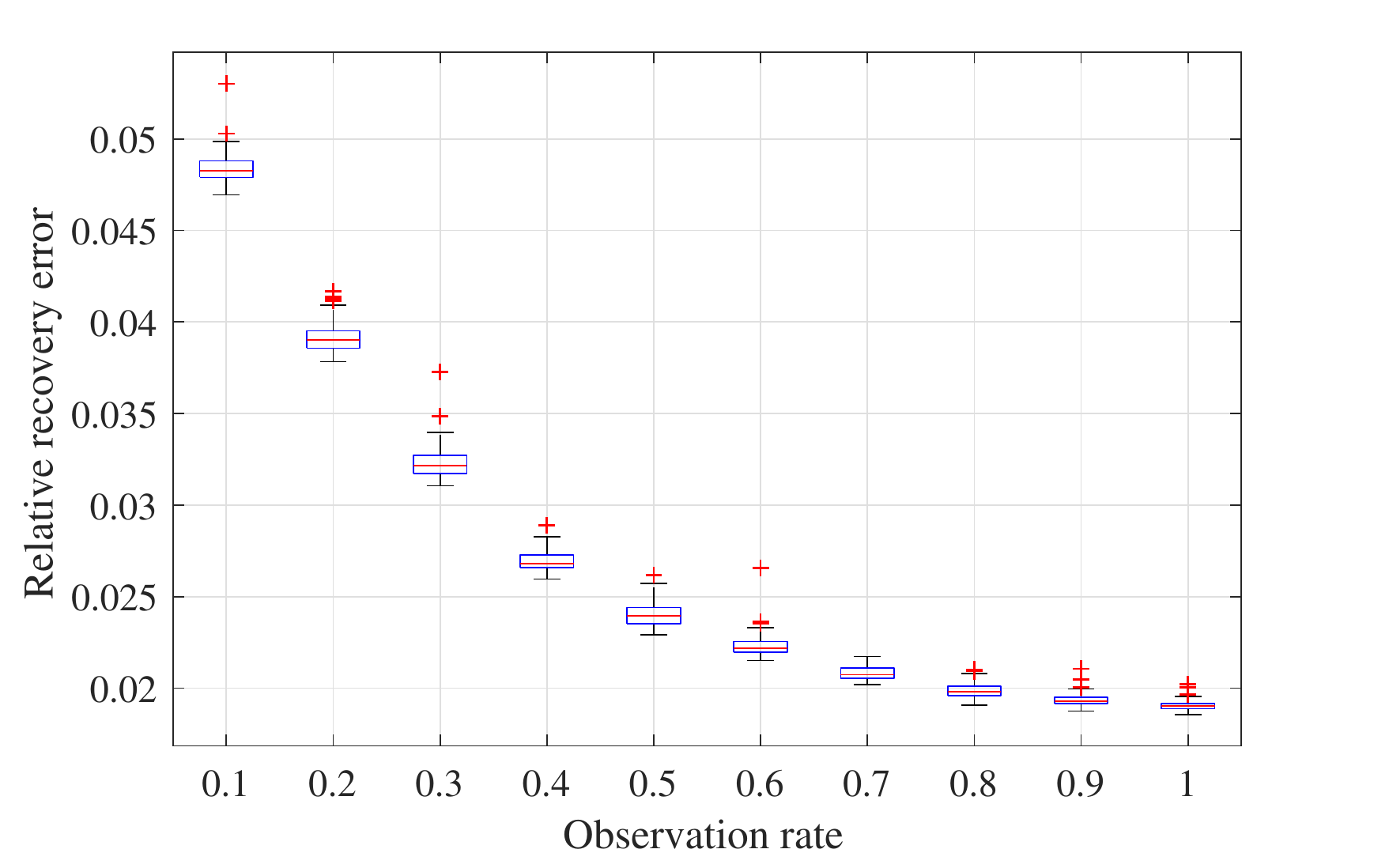}
	\caption{Relative recovery error when the observation rate changes ($W=3$, TAPGD)}
	\label{fig5}
\end{figure}

In Fig.~\ref{fig5}, we show a box-plot-diagram of relative recovery error with 100 runs obtained by TAPGD. All the setups are the same as the scenario $W=3$ in Fig.~\ref{fig4} (a). The tops and bottoms of each "box" are the 25th and 75th percentiles of the samples respectively. The maximum standard deviation happens when the observation rate is 0.3, which equals to $8.79\times 10^{-4}$. The relative standard deviation, which is defined as the ratio of the standard deviation to the mean, reaches its maximum value 0.028 when the observation rate is 0.6.

\begin{figure}[h]
	\centering
	\includegraphics[trim=5 5 10 10,clip,width=0.75\linewidth]{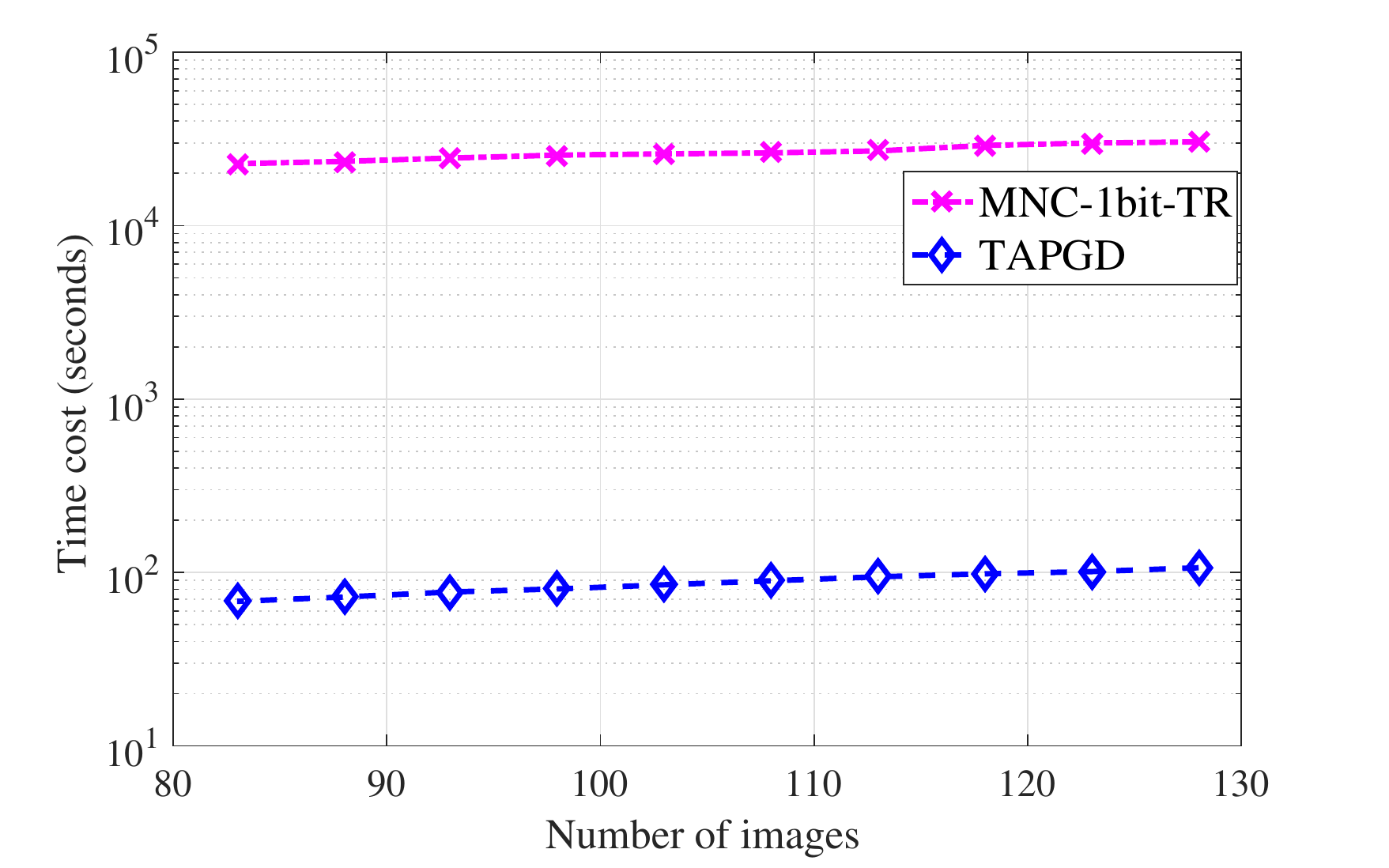}
	\caption{Time cost of TAPGD and MNC-1bit-TR}
	\label{fig6}
\end{figure}

\begin{figure}[h]
	\centering
	\includegraphics[width=0.8\linewidth]{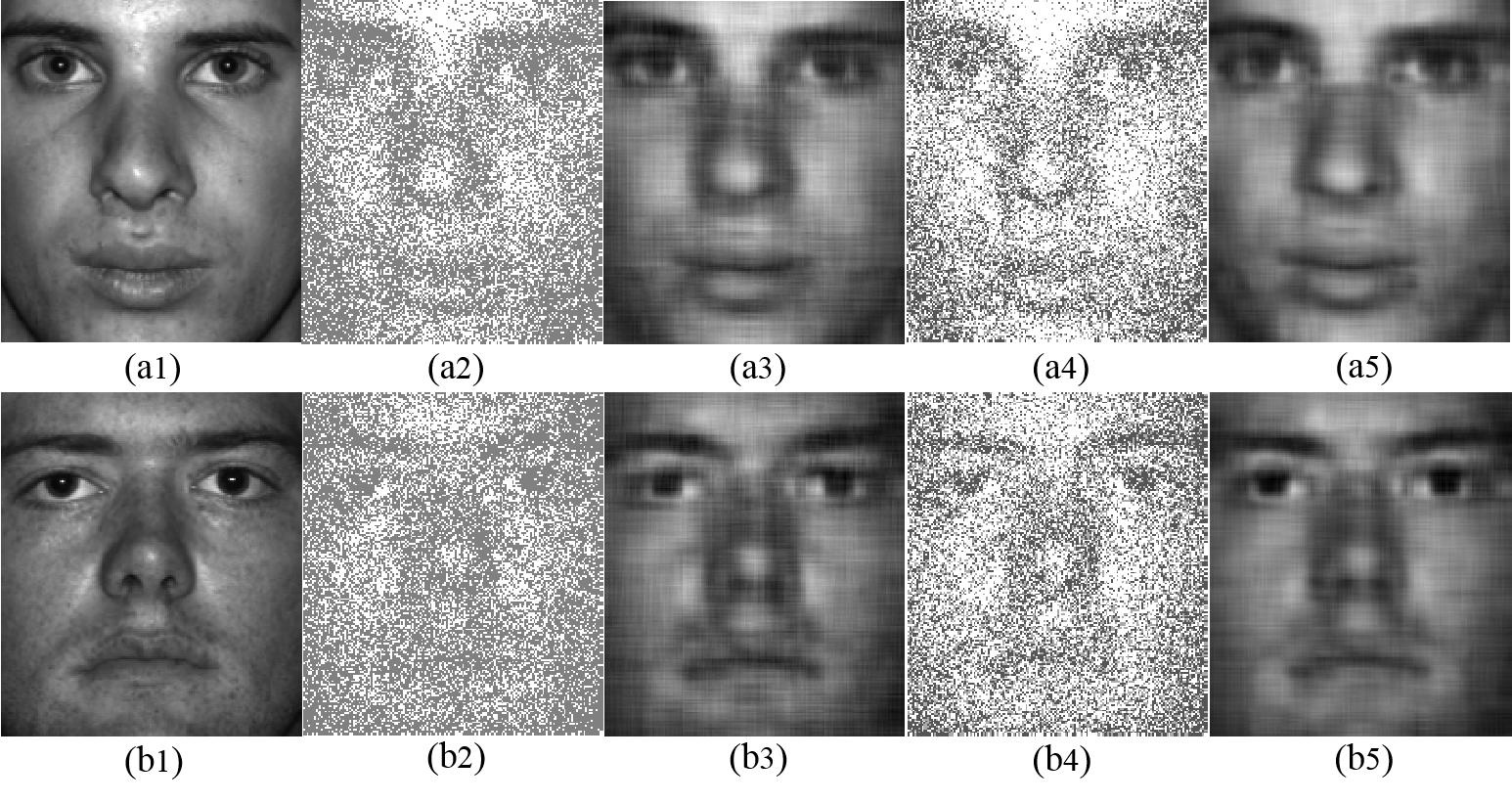}
	\caption{(a1,b1) Original images (a2,b2) Quantized images ($W=2$) (a3,b3) Recovered images ($W=2$) (a4,b4) Quantized images ($W=3$) (a5,b5) Recovered images ($W=3$)}
	\label{fig7}
\end{figure}

Fig.~\ref{fig6} compares the time cost of TAPGD and MNC-1bit-TR \cite{GPY19} when the number of facial images changes. TAPGD is three magnitudes faster than MNC-1bit-TR. Fig.~\ref{fig7} visualizes the quantized and recovered images by TAPGD.


\subsection{In-car music recommender dataset}
\begin{figure}[H]
	\centering
	\includegraphics[width=0.75\linewidth]{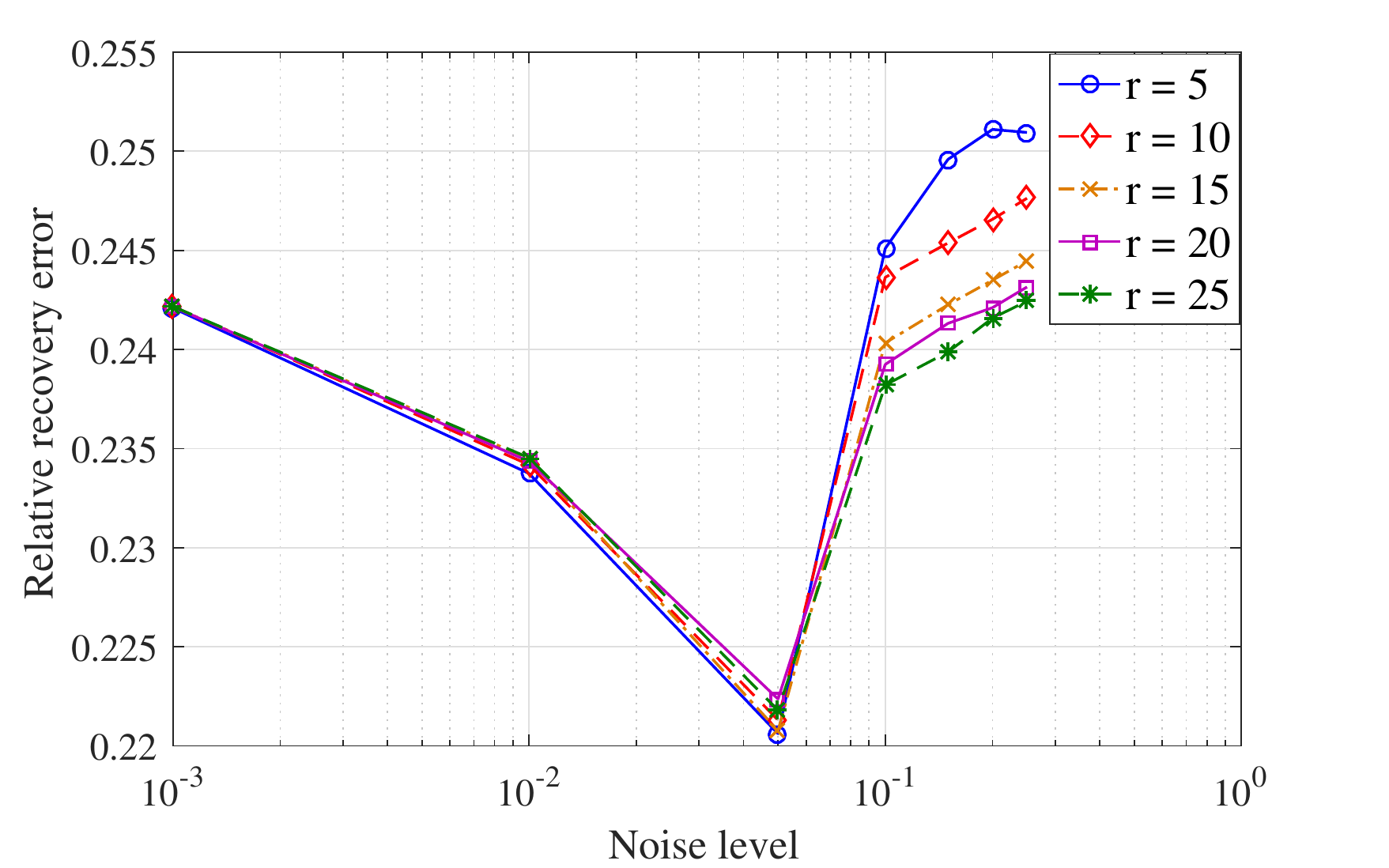}
	\caption{Relative recovery error when the estimated noise level and rank change}
	\label{fig8}
\end{figure}

Many recommender systems' ratings from users are highly quantized (such as like or dislike) with many missing entries (e.g., users don't give rating for all subjects), while the underlying systems may want to recover real-valued user ratings. Our method can be used to recover the true underlying real-valued user preferences, thus improving the quality of recommendations. We apply our method to an in-car music recommender dataset \cite{BLM11}. The recommender dataset contains 139 songs with 4012 ratings from 42 users. This dataset has 26 contexts that including relaxed driving, country side, happy, sleepy, etc. The same user may rate different scores to the same song under different contexts. 2751 ratings have the corresponding context information while the rest 1261 ratings do not have context information. We only use the ratings with context information. An example of three ratings is shown in Table \ref{table1}.
\begin{table}[h]
	\center
	\caption{Example of the in-car music recommender dataset \protect\cite{BLM11}}
	\label{table1}
	\begin{tabular}{l|l|l|l}
		UserID & ItemID & Rating & DrivingStyle \\
		6 & 1 & 4 & NA \\
		11 & 1 & 4 & NA\\ 
    	9 & 42 & 1 & NA\\ 
	\end{tabular}
		\begin{tabular}{l|l|l|l}
		Landscape & Mood & NaturalPhenomena & RoadType \\
		NA & NA & NA & NA\\
		NA & happy & NA & NA\\ 
    	NA & NA & NA & NA\\ 
	\end{tabular}
	\begin{tabular}{l|l|l}
	Sleepiness & TrafficCondition & Weather \\
	free road & NA & NA \\
	NA & NA & NA \\ 
	NA & NA & sunny \\ 
\end{tabular}
\end{table}
We construct the resulting tensor $\mathcal{M}$ as $\{\text{users} \times \text{musics} \times \text{contexts}\}$, which is a $42 \times 139 \times 26$ tensor. The ratings are quantized to $0,1,2,3,4,5$. All the locations without ratings are set to be zero. The observation rate is $1.81\%$ in the tensor. We then randomly set $0.362\%$ of the data ($20\%$ of the observed data) to be zero and let $\Omega_{\text{predict}}$ denote the set of the indices. We predict data with indices belong to $\Omega_{\text{predict}}$ using the rest $1.448\%$ of the data ($80\%$ of the observed data), which are observed. In this test, we define the relative recovery error as 
\begin{equation}\label{error_define}
\frac{1}{|\Omega_{\text{predict}}|}\sum_{(i_1,i_2,i_3) \in \Omega_{\text{predict}}}\frac{|\mathcal{M}_{i_1,i_2,i_3}-\bar{\mathcal{M}}_{i_1,i_2,i_3}|}{5},
\end{equation}
where $\tilde{\mathcal{M}}$ is our estimation of the ground truth, and $\bar{\mathcal{M}}$ maps the values in $\tilde{\mathcal{M}}$ to their nearest quantized values. The reason for the occurrence of 5 at denominator is that the maximum difference between $\bar{\mathcal{M}}$ and $\mathcal{M}$ is 5. The error increases when the difference increases. Ref. \cite{GPY19} also studies on the same dataset and first maps the multi-level quantized values to binary values. It then deletes some binary values and evaluates the recovery error. The smallest recovery error is $0.23$ by their method. We remark that the multi-level prediction is harder than binary prediction in this application, since the binary case is to choose one out of two numbers, while the multi-level case is to choose one out of $W>2$ numbers. Here we estimate the rank $r$ and the noise level $\sigma$, since we do not know the actual rank and noise. In Algorithm 1, we choose the estimated rank $r$ from the set $\{5,10,15,20,25\}$, and choose the estimated standard variation $\sigma$ from the set $\{0.001,0.01,0.05,0.1,0.15,0.2,0.25\}$. The recovery results are shown in Fig.~\ref{fig8}. The relative recovery error reaches its smallest value when $r=5$ and $\sigma = 0.05$, and the smallest relative recovery error is $0.22$.

\section{Conclusion and discussion}\label{conclusion}
This paper recovers a low-rank tensor from quantized measurements. A constrained maximum log-likelihood problem is proposed to estimate the ground truth tensor. The recovery error is proved to be at most $O(\frac{r\sqrt{K\log(K)}}{\sqrt{n^{K-1}}})$ when boundaries are known. This error bound is significantly smaller than the errors bounds of the existing methods for one-bit tensor recovery and low-rank matrix recovery from quantized measurements. We also propose an efficient method Tensor-Based Alternating Proximal Gradient Descent (TAPGD) to solve the nonconvex optimization problem. TAPGD is guaranteed to converge to a critical point from any initial point. TAPGD handles missing data and does not require information of the quantization rule. The method is evaluated on synthetic data, the Extend Yale Face Dataset B, and the in-car music recommender dataset. Future works include data recovery when partial measurements contain significant errors.

\section*{Appendix}\label{sec:appendix}
\subsection{Supporting lemmas used in the proof of Theorem 1}
Let $\langle \mathcal{A}, \mathcal{B} \rangle$ denote the inner product of $\mathcal{A}  \in \mathbb{R}^{n_1 \times ... \times n_K}$ and $\mathcal{B}  \in \mathbb{R}^{n_1 \times ... \times n_K}$, i.e., the sum of the products of their entries. Then the spectral norm of a tensor $\mathcal{X} \in \mathbb{R}^{n_1 \times ... \times n_K}$ is defined as
\begin{equation}
\begin{aligned}
&\|\mathcal{X}\| \\& = \text{sup}\{\langle \mathcal{X}, u_1 \circ u_2 ... \circ u_K \rangle: \|u_k\|_2=1, \\&~~~~~ u_k \in \mathbb{R}^{n_k}, \forall k \in [K]\} \\&= \text{sup}_{u_1,u_2,...,u_K} \mathcal{X}(u_1,u_2,...,u_K), \|u_k\|_2=1, \\&~~~~~ u_k \in \mathbb{R}^{n_k}, \forall k \in [K]
\\& = \text{sup}_{u_1,u_2,...,u_K} \sum_{i_1,i_2,...,i_K} \mathcal{X}_{i_1,i_2,...,i_K}u_{1i_1}u_{2i_2},...,u_{Ki_K},\\&~~~~~ \|u_k\|_2=1, u_k \in \mathbb{R}^{n_k}, \forall k \in [K]
\end{aligned}
\end{equation}
where $u_1 \circ u_2 ... \circ u_K \in \mathbb{R}^{n_1 \times ... \times n_K}$.

Lemma \ref{lemma1} provides an upper bound on the spectral norm of a tensor with independent random entries.
\begin{lemma}\label{lemma1}
	Suppose that $\mathcal{X} = [\mathcal{X}_{i_1,i_2,...,i_K}]_{1\le i_1\le n_1,1\le i_2 \le n_2,...,1 \le i_K \le n_K}$ is a $K$-dimensional tensor whose entries are independent random variables that satisfy, for some $s^2$,
	\begin{equation}
	\mathbb{E}[\mathcal{X}_{i_1,i_2,...,i_K}] = 0, \ \ \mathbb{E}[e^{\epsilon\mathcal{X}_{i_1,i_2,...,i_K}}]\le e^{s^2\epsilon^2/2}, \ \ a.s.
	\end{equation}
	Then
	\begin{equation}
	P(\|\mathcal{X}\| \ge \mu) \le \delta
	\end{equation}
	for some $\delta \in [0,1]$, where 
	\begin{equation}
	\mu = \sqrt{8s^2((\sum_{k=1}^{K}n_k)\log(4K/3)+\log(2/\delta))}.
	\end{equation}

\end{lemma}

\begin{IEEEproof}
The proof is completed by combining Lemma 1 and Theorem 1 in \cite{TS14}.
\end{IEEEproof}
We first define $F(\mathcal{X})$ as the function when $F_\Omega(\mathcal{X},\omega_1,\omega_2,\cdots,\omega_{W-1})$ is under the full observation and $\omega_l, \forall l \in [W-1]$ are known. Specifically,
\begin{equation}\label{def_main}
\begin{aligned}
&F(\mathcal{X}) \\&= -\sum_{i_1=1}^{n_1}\sum_{i_2=1}^{n_2}\cdots \sum_{i_K=1}^{n_K} \sum_{l=1}^W \\&~~~~\boldsymbol{1}_{[\mathcal{Y}_{i_1,i_2,...,i_K}=l]}\log(f_l(\mathcal{X}_{i_1,i_2,...,i_K},\omega_{l-1}^*, \omega_l^*)).
\end{aligned}
\end{equation}

Lemma \ref{lemma2} and Lemma \ref{lemma3} describe the relation of $\mathcal{X}^*$ with any data in the feasible set $\mathcal{S}_f$. Considering any $\mathcal{X}' \in \mathcal{S}_f$, we can calculate the second-order Tayler expansion of $F(\mathcal{X}')$ at $\mathcal{X}^*$. Lemma \ref{lemma2} indicates that the absolute value of the first-order term of the Taylor expansion can always be upper bounded by a term related to $\|\mathcal{X}'-\mathcal{X}^*\|_F$.
\begin{lemma}\label{lemma2}
	Let $\theta'=\mathrm{vec}(\mathcal{X}')$, $\theta^*=\mathrm{vec}(\mathcal{X}^*)$, $\nabla_{\theta}F(\theta^*)=\mathrm{vec}(\nabla_{\mathcal{X}}F(\mathcal{X}^*))$, and $\mathcal{X}'$, $\mathcal{X}^* \in \mathcal{S}_f$. Then with probability at least $1-\delta$, 
	\begin{equation}
	\begin{aligned}
	& |\left \langle \nabla_{\theta}F(\theta^*),\theta'-\theta^* \right \rangle| \le \\&2r\sqrt{8L_\alpha^2((\sum_{k=1}^{K}n_k)\log(4K/3)+\log(2/\delta))}\|\mathcal{X}'-\mathcal{X}^*\|_F,
	\end{aligned}
	\end{equation}
\end{lemma}

\begin{IEEEproof}
Consider 
\begin{equation*}
\begin{aligned}
\mathcal{Z}_{i_1,i_2,...,i_K}&:=[\nabla_{\mathcal{X}}F(\mathcal{X}^*)]_{i_1,i_2,...,i_K} \\&= -\sum_{l=1}^W\frac{\dot{f}_l(\mathcal{X}^*_{i_1,i_2,...,i_K})}{f_l(\mathcal{X}^*_{i_1,i_2,...,i_K})}\boldsymbol{1}_{[Y_{i_1,i_2,...,i_K}=l]}.
\end{aligned}
\end{equation*}
Recall that the probability $\mathcal{Y}_{i_1,i_2,\dots,i_K} = l$ given $\mathcal{X}_{i_1,i_2,\dots,i_K}^*$ is expressed by $f_l(\mathcal{X}_{i_1,i_2,\dots,i_K}^*)$, which only holds true for $X^*$. Then using the fact that $\sum_{l=1}^W f_l(\mathcal{X}_{i_1,i_2,...,i_K})=1$, we have $\mathbb{E}[\mathcal{Z}_{i_1,i_2,...,i_K}]=0$, $-L_{\alpha} \le \mathcal{Z}_{i_1,i_2,...,i_K}\le L_{\alpha}$. By Hoeffding's lemma, we can obtain $\mathbb{E}[e^{\epsilon Z^2_{i_1,i_2,...,i_K}}]\le e^{(L_\alpha + L_\alpha) ^2 \epsilon^2/8} = e^{L_\alpha^2 \epsilon^2/2}$. Replacing $s$ with $L_\alpha$ in Lemma \ref{lemma1}, we have
\begin{equation}
\begin{aligned}
&\|\nabla_{X}F(\mathcal{X}^*)\| \\&\le \sqrt{8L_\alpha^2((\sum_{k=1}^{K}n_k)\log(4K/3)+\log(2/\delta))}
\end{aligned}
\end{equation}
holds with probability at least $1-\delta$. 
Note that for $\forall \mathcal{X} \in \mathcal{S}_f$,
\begin{equation}\label{nuc_pro}
\begin{aligned}
&\|\mathcal{X}\|_* \\&\le \min \{\sqrt{\min(r_2, r_3,\dots,r_K)}\|\mathbf{X}_{(1)}\|_*,\\& \sqrt{\min(r_1, r_3,\dots,r_K)}\|\mathbf{X}_{(2)}\|_*,\\&\dots,\sqrt{\min(r_1, r_2,\dots,r_{K-1})}\|\mathbf{X}_{(K)}\|_*\}.
\end{aligned}
\end{equation}
where $r_k, k \in [K]$ is the $k$-rank of the tensor $\mathcal{X}$, which is defined as the column rank of $\mathbf{X}_{(k)}$. The tensor nuclear norm $\|\mathcal{X}\|_*$ is defined as
\begin{equation}
\begin{aligned}
\|\mathcal{X}\|_* = \text{inf}&\{\sum_{i=1}^{r}|\lambda_i|: \mathcal{X} = \sum_{i=1}^{r}\lambda_i v_{1,i} \circ v_{2,i} ... \circ v_{K,i},\\& \|v_{k,i}\|_2=1 \}
\end{aligned}
\end{equation}

The details of the property (\ref{nuc_pro}) can be viewed in Theorem 9.4 of \cite{FL18}. Note that $r_k \le r, \forall k \in [K]$, since $\mathbf{X}_{(k)} = \mathbf{A_k}(\mathbf{A_K}\odot \dots \odot \mathbf{A_{k+1}}\odot \mathbf{A_{k-1}}\odot \dots \mathbf{A_1})^T$. Therefore,

\begin{equation}
\begin{aligned}
&\|\mathcal{X}'-\mathcal{X}^*\|_* \\&\le \sqrt{2r}\|\mathbf{X}'_{(k)}-\mathbf{X}^*_{(k)}\|_* \\& \le 2r\|\mathbf{X}'_{(k)}-\mathbf{X}^*_{(k)}\|_F \\&= 2r\|\mathcal{X}'-\mathcal{X}^*\|_F.
\end{aligned}
\end{equation}
where the last inequality holds because $\|\cdot\|_* \le \sqrt{r}\|\cdot\|_F$ for any matrix.
We then have
\begin{equation}\label{part1}
\begin{aligned}
&\|\nabla_{\mathcal{X}}F(\mathcal{X}^*)\|\|\mathcal{X}'-\mathcal{X}^*\|_* \le \\& 2r\sqrt{8L_\alpha^2((\sum_{k=1}^{K}n_k)\log(4K/3)+\log(2/\delta))}\|\mathcal{X}'-\mathcal{X}^*\|_F
\end{aligned}
\end{equation}
holds with probability at least $1-\delta$.

Then,
\begin{equation*}
\begin{aligned}
& |\left \langle \nabla_{\theta}\mathcal{F}(\theta^*),\theta'-\theta^* \right \rangle| \\&= |\left \langle \nabla_{\mathcal{X}}F(\mathcal{X}^*),\mathcal{X}'-\mathcal{X}^* \right \rangle| \nonumber \\
& \le |\left \langle \nabla_{\mathcal{X}}F(\mathcal{X}^*),\mathcal{X}'-\mathcal{X}^* \right \rangle| \\
& \le \|\nabla_{\mathcal{X}}F(\mathcal{X}^*)\|\|\mathcal{X}'-\mathcal{X}^*\|_*\\
& \le 2r\sqrt{8L_\alpha^2((\sum_{k=1}^{K}n_k)\log(4K/3)+\log(\frac{2}{\delta}))}\|\mathcal{X}'-\mathcal{X}^*\|_F
\end{aligned}
\end{equation*}
holds with probability at least $1-\delta$. The second inequality comes from the fact $|\left \langle \mathcal{A}, \mathcal{B} \right \rangle| \le \|\mathcal{A}\|\|\mathcal{B}\|_*$ for two tensors $\mathcal{A}$ and $\mathcal{B}$. We then have the desired result.
\end{IEEEproof}
Lemma \ref{lemma3} provides a lower bound on the second-order term of the second-order Taylor expansion, which is also related to $\|\mathcal{X}'-\mathcal{X}^*\|_F$.
\begin{lemma}\label{lemma3}
	Let $\theta'=\mathrm{vec}(\mathcal{X}')$, $\theta^*=\mathrm{vec}(\mathcal{X}^*)$, and $\mathcal{X}'$, $\mathcal{X}^* \in \mathcal{S}_f$. Then for any $\tilde{\theta}=\theta^*+\eta(\theta'-\theta^*)$ and any $\eta \in [0,1]$, we have
	\begin{equation}
	\left \langle \theta'-\theta^*, (\nabla^2_{\theta\theta}F(\tilde{\theta}))(\theta'-\theta^*)\right \rangle \ge \gamma_{\alpha}\|\mathcal{X}'-\mathcal{X}^*\|_F^2.
	\end{equation}
\end{lemma}

\begin{IEEEproof}
Lemma \ref{lemma3} is an extension of Lemma 7 in \cite{Bhaskar16}.

Using (\ref{def_main}), it follows that
\begin{equation*}
\begin{aligned}
&\frac{\partial^2 F(\mathcal{X})}{\partial^2 \mathcal{X}_{i_1,i_2,...,i_K}} \\&= \sum_{l=1}^W(\frac{\dot{f}_l^2(\mathcal{X}_{i_1,i_2,...,i_K})}{f_l^2(\mathcal{X}_{i_1,i_2,...,i_K})}-\frac{\ddot{f}_l(\mathcal{X}_{i_1,i_2,...,i_K})}{f_l(\mathcal{X}_{i_1,i_2,...,i_K})})\boldsymbol{1}_{[\mathcal{Y}_{i_1,i_2,...,i_K}=l]}
\end{aligned}
\end{equation*}

Then we have
\begin{equation*}
\begin{aligned}
&\left \langle \theta'-\theta^*, (\nabla^2_{\theta\theta}F(\tilde{\theta}))(\theta'-\theta^*)\right \rangle \\&= {\sum_{i_1=1}^{n_1}}\cdots{\sum_{i_K=1}^{n_K}}(\frac{\partial^2 F(\tilde{\mathcal{X}})}{\partial^2 \mathcal{X}_{i_1,i_2,...,i_K}})(\mathcal{X}'_{i_1,i_2,...,i_K} - \mathcal{X}^*_{i_1,i_2,...,i_K}) \\&\ge \gamma_\alpha {\sum_{i_1=1}^{n_1}}\cdots{\sum_{i_K=1}^{n_K}} (\mathcal{X}'_{i_1,i_2,...,i_K} - \mathcal{X}_{i_1,i_2,...,i_K}^*)^2
\\& = \gamma_\alpha \|\mathcal{X}' - \mathcal{X}^*\|_F^2
\end{aligned}
\end{equation*}
where the first inequality comes from the fact that $\gamma_{\alpha} = \min_{l\in[W]}\inf_{|x|\le2\alpha}\{\frac{\dot{f}_l^2(x)}{f_l^2(x)}-\frac{\ddot{f}_l(x)}{f_l(x)}\}$.
\end{IEEEproof}
\subsection{Proof of Theorem 1}
\begin{IEEEproof}
The first bound follows from the fact that $\mathcal{X}',\mathcal{X}^* \in \mathcal{S}_f$. We have
\begin{equation}\label{first_bound}
\begin{aligned}
&\|\hat{\mathcal{X}}-\mathcal{X}^*\|_F/\sqrt{n_1n_2...n_K} \\& \le 2\alpha\sqrt{n_1n_2...n_K}/\sqrt{n_1n_2...n_K} \\&= 2\alpha.
\end{aligned}
\end{equation}
Let $\hat{\theta}=\mathrm{vec}(\hat{\mathcal{X}})$ and $\mathcal{F}(\hat{\theta})=F(\hat{\mathcal{X}})$. By the second-order Taylor's theorem, we have
\begin{equation}\label{taylor}
\begin{aligned}
F(\hat{\theta}) = &F(\theta^*) + \left \langle \nabla_{\theta}F(\theta^*),\hat{\theta}-\theta^* \right \rangle\\& + \frac{1}{2}\left \langle \theta-\theta^*, (\nabla^2_{\theta\theta}F(\tilde{\theta}))(\hat{\theta}-\theta^*) \right \rangle,
\end{aligned}
\end{equation}
where $\tilde{\theta}=\theta^*+\eta(\hat{\theta}-\theta^*)$ for some $\eta\in[0,1]$, with the corresponding tensor $\tilde{\mathcal{X}}=\mathcal{X}^*+\eta(\hat{\mathcal{X}}-\mathcal{X}^*)$.

Using the results of Lemma 2 and Lemma 3, we can obtain that
\begin{equation}
\begin{aligned}
& F(\hat{\mathcal{X}}) \ge F(\mathcal{X}^*)\\&-2r\sqrt{8L_\alpha^2((\sum_{k=1}^{K}n_k)\log(4K/3)+\log(2/\delta))}\|\hat{\mathcal{X}}-\mathcal{X}^*\|_F \\&+ \frac{\gamma_{\alpha}}{2}\|\hat{\mathcal{X}}-\mathcal{X}^*\|_F^2
\end{aligned}
\end{equation}
holds with probability at least $1-\delta$. Note that $\hat{\mathcal{X}}$ is the global optimal of the optimization problem. Thus, $F(\hat{\mathcal{X}}) \le F(\mathcal{X}^*)$. We then have
\begin{equation}
\begin{aligned}
& \frac{\gamma_{\alpha}}{2}\|\hat{\mathcal{X}}-\mathcal{X}^*\|_F^2 \\& \le 2r\sqrt{8L_\alpha^2((\sum_{k=1}^{K}n_k)\log(4K/3)+\log(2/\delta))}\|\hat{\mathcal{X}}-\mathcal{X}^*\|_F
\end{aligned}
\end{equation}
holds with probability at least $1-\delta$. Thus,
\begin{equation}\label{sec_bound}
\|\hat{\mathcal{X}}-\mathcal{X}^*\|_F/\sqrt{n_1n_2...n_K} \le U_{\alpha}.
\end{equation}
holds with the same probability $1-\delta$, where
\begin{equation}
\begin{aligned}
&U_{\alpha}  = \frac{4r\sqrt{8L_\alpha^2((\sum_{k=1}^{K}n_k)\log(4K/3)+\log(2/\delta))}}{\gamma_\alpha \sqrt{n_1n_2...n_K}}.
\end{aligned}
\end{equation}

Combining (\ref{first_bound}) and (\ref{sec_bound}), we have
\begin{equation}\label{main_result}
\|\hat{\mathcal{X}}-\mathcal{X}^*\|_F/\sqrt{n_1n_2...n_K} \le \min (2\alpha,U_{\alpha}),
\end{equation}
and this complete the proof.
\end{IEEEproof}
\subsection{TAPGD: Proof of the Lipschitz differential property and calculation of Lipschitz constants}\label{Dif}
We provide the Lipschitz differential property of $H$ and compute the corresponding Lipschitz constants of its partial derivatives with respect to $\mathbf{A_k}\in \mathbb{R}^{n_k \times r}, \forall k \in [K]$, $\mathcal{X} \in \mathbb{R}^{n_1 \times n_2 \times \dots \times n_K}$, and $\omega_{l},\forall l \in [W-1]$. We call a function Lipschitz differentiable if and only if all its partial derivatives are Lipschitz continuous. The definition of Lipschitz continuous of a function's partial derivatives is shown in Definition \ref{Def1}.

\begin{defi}\cite{BST14}\label{Def1}
	For any variable $y$, and a function $y \rightarrow \Upsilon(y,z_1,z_2,...,z_n)$, with other variables $z_1, z_2,.., z_n$ fixed, the partial derivative $\nabla_{y} \Upsilon(y,z_1,z_2,...,z_n)$ is said to be Lipschitz continuous with Lipschitz constant $L_p(z_1,z_2,...,z_n)$, if the following relation holds
	\begin{equation*}
	\begin{aligned}
	&\| \nabla_{y} \Upsilon(y_1,z_1,z_2,...,z_n) - \nabla_{y} \Upsilon(y_2,z_1,z_2,...,z_n) \|_F \\& \le L_p(z_1,z_2,...,z_n) \| y_1 - y_2 \|_F,~~ \forall y_1,y_2.
	\end{aligned}
	\end{equation*}
\end{defi}
Let $L^{t+1}_{\mathbf{A_k}},\forall k \in [K]$, $L^{t+1}_{\mathcal{X}}$, and $L^{t+1}_{\omega_{l}}, \forall l \in [W-1]$ denote the smallest Lipschitz constants of $\nabla_{\mathbf{A_k}} H$, $\nabla_{\mathcal{X}} H$, and $\nabla_{\omega_{l}} H$ in the $(t+1)$-th iteration. The details of the calculation are shown in (\ref{L1}), (\ref{L2}), and (\ref{L3}).

\begin{equation}\label{L1}
\begin{aligned}
&\| \nabla_{\mathbf{A_k}} H(\mathbf{A_k}) - \nabla_{\mathbf{A_k}} H(\mathbf{A_k}') \|_F \\ & = \| (\mathbf{A_k}(\mathbf{B_k}^t)^T - \mathcal{X}_{(k)}) \mathbf{B_k}^t \\&~~~~~ - (\mathbf{A_k}'(\mathbf{B_k}^t)^T - \mathcal{X}_{(k)}) \mathbf{B_k}^t\|_F
 \\&= \| (\mathbf{A_k}-\mathbf{A_k}')(\mathbf{B_k}^t)^T \mathbf{B_k}^t\|_F
\\& \stackrel{(\rm{a})} \le \|(\mathbf{B_k}^t)^T \mathbf{B_k}^t\|\|\mathbf{A_k}-\mathbf{A_k}'\|_F
\\&\stackrel{(\rm{b})} = \frac{1}{\tau_{\mathbf{A_k}}(\mathbf{B_k}^t)} \| \mathbf{A_k}-\mathbf{A_k}' \|_F,
\end{aligned}
\end{equation}
where $\nabla_{\mathbf{A_k}} H(\mathbf{A_k})$ and $\nabla_{\mathbf{A_k}} H(\mathbf{A_k}')$ are the abbreviations of $\nabla_{\mathbf{A_k}} H(\mathbf{A_1}^{t+1},\- \mathbf{A_2}^{t+1},\- \cdots,\- \mathbf{A_{k-1}}^{t+1},\- \mathbf{A_k},\- \mathbf{A_{k+1}}^t,\- \cdots,\- \mathbf{A_K}^t,\- \mathcal{X}^t,\- \omega_{1}^t,\- \omega_{2}^t,\- \cdots,\- \omega_{W-1}^t)$ and $\nabla_{\mathbf{A_k}} H(\mathbf{A_1}^{t+1},\- \mathbf{A_2}^{t+1},\- \cdots,\- \mathbf{A_{k-1}}^{t+1},\- \mathbf{A_k}',\- \mathbf{A_{k+1}}^t,\- \cdots,\- \mathbf{A_K}^t,\- \mathcal{X}^t,\- \omega_{1}^t,\- \omega_{2}^t,\- \cdots,\- \omega_{W-1}^t)$, respectively. $\mathbf{B_k}^t$ represents $\mathbf{A_K}^t\odot ...\odot \mathbf{A_{k+1}}^t \odot \mathbf{A_{k-1}}^{t+1} \odot ...\odot \mathbf{A_1}^{t+1}$. (a) holds from the inequality $\|\mathbf{A}\mathbf{B}\|_F \le \|\mathbf{A}\|\|\mathbf{B}\|_F$. (b) follows from
\begin{equation}
\begin{aligned}
\tau_{\mathbf{A_k}} = \frac{1}{\|(\mathbf{B_k})^T\mathbf{B_k}\|},\forall k \in [K],
\end{aligned}
\end{equation}
and (\ref{L1}) implies that
\begin{equation}
\begin{aligned}
&L^{t+1}_{\mathbf{A_k}} \leq  \|(\mathbf{B_k}^t)^T \mathbf{B_k}^t\|, \textrm{ and }
\\&\tau_{\mathbf{A_k}}(\mathbf{B_k}^t)  \le 1/L^{t+1}_{\mathbf{A_k}}.
\end{aligned}
\end{equation}

\begin{equation}\label{L2}
\begin{aligned}
&\| \nabla_{\mathcal{X}} H(\mathcal{X}) - \nabla_{\mathcal{X}} H(\mathcal{X}') \|_F \\ & = \| \nabla_\mathcal{X} F_\Omega(\mathcal{X}, \omega_{1}^t, \omega_{2}^t, \cdots, \omega_{W-1}^t) \\&~~~~~ + \lambda (\mathcal{X} - \mathbf{A_1}^{t+1} \circ \mathbf{A_2}^{t+1} \circ...\circ \mathbf{A_K}^{t+1}) \\&~~~~~- \nabla_\mathcal{X} F_\Omega(\mathcal{X}', \omega_{1}^t, \omega_{2}^t, \cdots, \omega_{W-1}^t) \\&~~~~~- \lambda (\mathcal{X}' - \mathbf{A_1}^{t+1} \circ \mathbf{A_2}^{t+1} \circ...\circ \mathbf{A_K}^{t+1})\|_F
\\& \stackrel{(\rm{c})} = \| \nabla_\mathcal{X} F_\Omega(\mathcal{X}, \omega_{1}^t, \omega_{2}^t, \cdots, \omega_{W-1}^t) \\&~~~~~- \nabla_\mathcal{X} F_\Omega(\mathcal{X}', \omega_{1}^t, \omega_{2}^t, \cdots, \omega_{W-1}^t) \|_F + \|\lambda(\mathcal{X}-\mathcal{X}')\|_F
\\& \stackrel{(\rm{d})} = \| \mathrm{diag}(\nabla^2 F_\Omega(\bar{\mathcal{X}}))\mathrm{vec}(\mathcal{X}-\mathcal{X}') \|_2 + \|\lambda(\mathcal{X}-\mathcal{X}')\|_F
 \\&\stackrel{(\rm{e})} = (\| \mathrm{diag}(\nabla^2 F_\Omega(\bar{\mathcal{X}}))\|_\infty  + \lambda) \|\mathcal{X}-\mathcal{X}'\|_F
\\& \stackrel{(\rm{f})} \le (\frac{1}{\sigma^2\beta^2} + \lambda) \|\mathcal{X}-\mathcal{X}'\|_F
 \\&\stackrel{(\rm{g})} = \frac{1}{\tau_{\mathcal{X}}} \| \mathcal{X}-\mathcal{X}'\|_F,
\end{aligned}
\end{equation}
Where $\nabla_{\mathcal{X}} H(\mathcal{X})$ and $\nabla_{\mathcal{X}} H(\mathcal{X}')$ are the abbreviations of $\nabla_{\mathcal{X}} H(\mathbf{A_1}^{t+1},\- \mathbf{A_2}^{t+1},\- \cdots,\- \mathbf{A_K}^{t+1},\- \mathcal{X},\- \omega_{1}^t,\- \omega_{2}^t,\- \cdots,\- \omega_{W-1}^t)$ and $\nabla_{\mathcal{X}} H(\mathbf{A_1}^{t+1},\- \mathbf{A_2}^{t+1},\- \cdots,\- \mathbf{A_K}^{t+1},\- \mathcal{X}',\- \omega_{1}^t,\- \omega_{2}^t,\- \cdots,\- \omega_{W-1}^t)$, respectively. In (\ref{L2}), (c) comes from the triangle inequality. (d) follows from the differential mean value theorem, and the fact $\|\mathbf{A}\|_F = \|\mathrm{vec}(\mathbf{A})\|_2$. $\nabla^2 F_\Omega(\bar{\mathcal{X}}) \in \mathbb{R}^{n_1 \times n_2 \times \dots \times n_K}$ has the $(i_1,i_2, \dots,i_K)$-th entry equaling to ${\frac{\partial^2 F_\Omega}{\partial^2 \mathcal{X}_{i_1,i_2, \dots,i_K}}|}_{\bar{\mathcal{X}}_{i_1,i_2, \dots,i_K}}$, and $\mathrm{diag}(\nabla^2 F_\Omega(\bar{\mathcal{X}})) \in \mathbb{R}^{n_1 n_2 \dots n_K\times n_1 n_2 \dots n_K}$ is a  diagonal matrix with the diagonal vector equaling to $\mathrm{vec}(\nabla^2 F_\Omega(\bar{\mathcal{X}}))$. (e) follows from the fact that the $l_2$ norm of a diagonal matrix is equal to its entrywise infinity norm. Note that the probability distribution function of the normal distribution and its derivative have the upper bounds  $\frac{1}{\sqrt{2\pi} \sigma}$ and $\frac{e^{-1/2}}{\sqrt{2\pi} \sigma^2}$, respectively. Then one can check that $\|\mathrm{diag}(\nabla^2 F_\Omega(\bar{\mathcal{X}}))\|_\infty$ is bounded by $\frac{1}{\sigma^2\beta^2}$. (f) follows from upper bounding $\|\mathrm{diag}(\nabla^2 F_\Omega(\bar{\mathcal{X}}))\|_\infty$ with $\frac{1}{\sigma^2\beta^2}$. (g) comes from $\tau_{\mathcal{X}} = \frac{1}{\frac{1}{\sigma^2\beta^2}+\lambda}$. Therefore, $\tau_{\mathcal{X}}  \le 1/L^{t+1}_{\mathcal{X}}$.

\begin{equation}\label{L3}
\begin{aligned}
&\| \nabla_{\omega_{l}} H(\omega_{l}) - \nabla_{\omega_{l}} H(\omega_{l}') \|_F \\ & = \| \sum_{(i_1,i_2,\cdots,i_K) \in \Omega} \\&~~~~~(\frac{\boldsymbol{1}_{[\mathcal{Y}_{i_1,i_2,...,i_K}=l+1]}\dot{\Phi}(\omega_l-\mathcal{X}_{i_1,i_2,\dots,i_K}^{t+1})}{\Phi(\omega_{l+1}^t-\mathcal{X}_{i_1,i_2,\dots,i_K}^{t+1})-\Phi(\omega_{l}-\mathcal{X}_{i_1,i_2,\dots,i_K}^{t+1})} \\&~~~~~- \frac{\boldsymbol{1}_{[\mathcal{Y}_{i_1,i_2,...,i_K}=l]}\dot{\Phi}(\omega_l-\mathcal{X}_{i_1,i_2,\dots,i_K}^{t+1})}{\Phi(\omega_l-\mathcal{X}_{i_1,i_2,\dots,i_K}^{t+1})-\Phi(\omega_{l-1}^{t+1}-\mathcal{X}_{i_1,i_2,\dots,i_K}^{t+1})}) \\&~~~~~-\sum_{(i_1,i_2,\cdots,i_K) \in \Omega}\\&~~~~~ (\frac{\boldsymbol{1}_{[\mathcal{Y}_{i_1,i_2,...,i_K}=l+1]}\dot{\Phi}(\omega_l'-\mathcal{X}_{i_1,i_2,\dots,i_K}^{t+1})}{\Phi(\omega_{l+1}^t-\mathcal{X}_{i_1,i_2,\dots,i_K}^{t+1})-\Phi(\omega_{l}'-\mathcal{X}_{i_1,i_2,\dots,i_K}^{t+1})} \\&~~~~~- \frac{\boldsymbol{1}_{[\mathcal{Y}_{i_1,i_2,...,i_K}=l]}\dot{\Phi}(\omega_l'-\mathcal{X}_{i_1,i_2,\dots,i_K}^{t+1})}{\Phi(\omega_l'-\mathcal{X}_{i_1,i_2,\dots,i_K}^{t+1})-\Phi(\omega_{l-1}^{t+1}-\mathcal{X}_{i_1,i_2,\dots,i_K}^{t+1})})\|_F
\\& \stackrel{(\rm{h})} \le \| \langle \mathcal{G}_{l+1}, \nabla J(\mathcal{U}_{\omega_l}) \rangle (\omega_l-\omega_l')\|_F \\&~~~~~+ \|\langle \mathcal{G}_{l}, \nabla M(\mathcal{V}_{\omega_l}) \rangle (\omega_l-\omega_l')\|_F
\\& \stackrel{(\rm{i})} \le \|\mathcal{G}_{l+1}\|_F\|\nabla J(\mathcal{U}_{\omega_l})\|_\infty\|\omega_l-\omega_l'\|_F\\&~~~~~ + \|\mathcal{G}_{l}\|_F\|\nabla M(\mathcal{V}_{\omega_l})\|_\infty\|\omega_l-\omega_l'\|_F
\\& \stackrel{(\rm{j})} \le \|\mathcal{G}_{l+1}\|_F\frac{1}{\sigma^2\beta^2}\|\omega_l-\omega_l'\|_F + \|\mathcal{G}_{l}\|_F\frac{1}{\sigma^2\beta^2}\|\omega_l-\omega_l'\|_F
\\& = \frac{1}{\sigma^2\beta^2}(\sqrt{G_{l}}+\sqrt{G_{l+1}})\|\omega_l-\omega_l'\|_F
\\& \stackrel{(\rm{k})} = \frac{1}{\tau_{\omega_{l}}} \| \omega_l-\omega_l'\|_F,
\end{aligned}
\end{equation}
where $\nabla_{\omega_{l}} H(\omega_{l})$ andv$\nabla_{\omega_{l}} H(\omega_{l}')$ are the abbreviations of $\nabla_{\omega_{l}} H(\mathbf{A_1}^{t+1},\- \mathbf{A_2}^{t+1},\- \cdots,\- \mathbf{A_K}^{t+1},\- \mathcal{X}^{t+1},\- \omega_{1}^{t+1},\- \omega_{2}^{t+1},\- \cdots,\- \omega_{l-1}^{t+1},\- \omega_{l},\- \omega_{l+1}^t,\- \cdots,\- \omega_{W-1}^t)$ and $\nabla_{\omega_{l}} H(\mathbf{A_1}^{t+1},\- \mathbf{A_2}^{t+1},\- \cdots,\- \mathbf{A_K}^{t+1},\- \mathcal{X}^{t+1},\- \omega_{1}^{t+1},\- \omega_{2}^{t+1},\- \cdots,\- \omega_{l-1}^{t+1},\- \omega_{l}',\- \omega_{l+1}^t,\- \cdots,\- \omega_{W-1}^t)$, respectively. In (\ref{L3}), $\mathcal{G}_{l},\mathcal{G}_{l+1}$ are binary tensors with entries equaling to one when the corresponding positions of $\mathcal{Y}$ equal to $l$ and $l+1$, respectively, and with entries equaling to zero otherwise. (h) follows from the differential mean value theorem, and $\mathcal{U}_{\omega_l},\mathcal{V}_{\omega_l} \in \mathbb{R}^{n_1 \times n_2 \times \dots \times n_K}$ have the entries between $\omega_l$ and $\omega_l'$ to satisfy the differential mean value theorem. The $(i_1,i_2, \dots,i_K)$-th entries of $\nabla J(\mathcal{U}_{\omega_l}),\nabla M(\mathcal{V}_{\omega_l}) \in \mathbb{R}^{n_1 \times n_2 \times \dots \times n_K}$ are partial derivatives of $\frac{\dot{\Phi}(\omega_l-\mathcal{X}_{i_1,i_2,\dots,i_K}^{t+1})}{\Phi(\omega_{l+1}^t-\mathcal{X}_{i_1,i_2,\dots,i_K}^{t+1})-\Phi(\omega_{l}-\mathcal{X}_{i_1,i_2,\dots,i_K}^{t+1})}$, and $\frac{\dot{\Phi}(\omega_l-\mathcal{X}_{i_1,i_2,\dots,i_K}^{t+1})}{\Phi(\omega_l-\mathcal{X}_{i_1,i_2,\dots,i_K}^{t+1})-\Phi(\omega_{l-1}^{t+1}-\mathcal{X}_{i_1,i_2,\dots,i_K}^{t+1})}$ with respect to $\omega_l$ at the points $(\mathcal{U}_{\omega_l})_{i_1,i_2, \dots,i_K}$ and $(\mathcal{V}_{\omega_l})_{i_1,i_2, \dots,i_K}$, respectively. (j) comes from the fact that $\|\nabla J(\mathcal{U}_{\omega_l})\|_\infty$ and $\|\nabla M(\mathcal{V}_{\omega_l})\|_\infty$ are upper bounded by $\frac{1}{\sigma^2\beta^2}$. (k) comes from $\tau_{\omega_l} = \frac{\sigma^2\beta^2}{\sqrt{G_{l}}+\sqrt{G_{l+1}}},\forall l \in [W-1]$. Thus, $\tau_{\omega_{l}}  \le 1/L^{t+1}_{\omega_{l}}$.

We remark that the results of (\ref{L1}) and (\ref{L2}) do not change when the boundaries $\omega_{l}^*, \forall l \in [W-1]$ are known to TAPGD, since $\omega_{l}^{t=1}, \forall l \in [W-1]$ are fixed values in (\ref{L1}) and (\ref{L2}).

\subsection{Proof of Theorem 2}
\begin{IEEEproof}
As described in Section \ref{algorithm:tapgd} of the paper,
$\psi_1(\mathcal{X})$ corresponds to the operations of setting $\mathcal{X}_{i_1,i_2,...,i_K}$ to $\alpha$ if $\mathcal{X}_{i_1,i_2,...,i_K} > \alpha$, and setting $\mathcal{X}_{i_1,i_2,...,i_K}$ to $-\alpha$ if $\mathcal{X}_{i_1,i_2,...,i_K} < -\alpha$, $\forall i_k \in [n_k], k \in [K]$. $\psi_2(\omega_l)$ corresponds to the operations of setting $\omega_l = \min(\omega_{l+1} - \kappa_{l+1}, \alpha_{\text{upper}})$ if $\omega_l > \min(\omega_{l+1} - \kappa_{l+1}, \alpha_{\text{upper}})$, and setting $\omega_l = \max(\omega_{l-1} + \kappa_l, \alpha_{\text{low}})$ if $\omega_l < \max(\omega_{l-1} + \kappa_l, \alpha_{\text{low}})$, $\forall l \in [W-1]$. TAPGD is a special case of the Proximal Alternating Linearized Minimization (PALM) algorithm from the results in \cite{BST14}. The global convergence of TAPGD to a critical point of (12) from any initial point can be proved by two steps: (1) $H(\mathbf{A_1}, \mathbf{A_2}, \cdots, \mathbf{A_K}, \mathcal{X}, \omega_1, \omega_2,\cdots, \omega_{W-1})$ is Lipschitz differentiable; (2) $H(\mathbf{A_1}, \mathbf{A_2}, \cdots, \mathbf{A_K}, \mathcal{X}, \omega_1, \omega_2,\cdots, \omega_{W-1}) + \Psi_1(\mathcal{X}) + \sum_{l=1}^{W-1}\Psi_2(\omega_l)$ satisfies the Kurdyka-Lojasiewicz (KL) property.

The Lipschitz differential property of $H(\mathbf{A_1}, \mathbf{A_2}, \cdots, \mathbf{A_K}, \mathcal{X}, \omega_1, \omega_2,\cdots, \omega_{W-1})$ has been shown in Section \ref{Dif}. $\Psi_1$ and $\Psi_2$ are semi-algebraic functions. According to \cite{BST14}, a semi-algebraic function satisfies the KL property. In addition, function $H(\mathbf{A_1}, \mathbf{A_2}, \cdots, \mathbf{A_K}, \mathcal{X}, \omega_1, \omega_2,\cdots, \omega_{W-1})$ is differentiable everywhere, which is equivalent to being real analytic. Thus, $H(\mathbf{A_1}, \mathbf{A_2}, \cdots, \mathbf{A_K}, \mathcal{X}, \omega_1, \omega_2,\cdots, \omega_{W-1})$ is a KL function according to \cite{XY13}. Finally, we have $H(\mathbf{A_1}, \mathbf{A_2}, \cdots, \mathbf{A_K}, \mathcal{X}, \omega_1, \omega_2,\cdots, \omega_{W-1}) + \Psi_1(\mathcal{X}) + \sum_{l=1}^{W-1}\Psi_2(\omega_l)$ satisfying the KL property. The claim follows by \cite{XY13}. By Remark 3.4 in \cite{BST14}, the convergence rate is at least $O(t^{\frac{\theta - 1}{2\theta - 1}})$, for some $\theta \in (\frac{1}{2},1)$. The proof is done.
\end{IEEEproof}

\section*{Acknowledgment}

This research is supported in part by ARO W911NF-17-1-0407, NSF \#1508875 and the ERC Program of NSF and DoE under the supplement to NSF Award EEC-1041877 and the CURENT Industry Partnership Program.


\begin{thebibliography}{10}
	\providecommand{\url}[1]{#1}
	\csname url@samestyle\endcsname
	\providecommand{\newblock}{\relax}
	\providecommand{\bibinfo}[2]{#2}
	\providecommand{\BIBentrySTDinterwordspacing}{\spaceskip=0pt\relax}
	\providecommand{\BIBentryALTinterwordstretchfactor}{4}
	\providecommand{\BIBentryALTinterwordspacing}{\spaceskip=\fontdimen2\font plus
		\BIBentryALTinterwordstretchfactor\fontdimen3\font minus
		\fontdimen4\font\relax}
	\providecommand{\BIBforeignlanguage}[2]{{%
			\expandafter\ifx\csname l@#1\endcsname\relax
			\typeout{** WARNING: IEEEtranS.bst: No hyphenation pattern has been}%
			\typeout{** loaded for the language `#1'. Using the pattern for}%
			\typeout{** the default language instead.}%
			\else
			\language=\csname l@#1\endcsname
			\fi
			#2}}
	\providecommand{\BIBdecl}{\relax}
	\BIBdecl
	
	\bibitem{ATT18}
	A.~Aidini, G.~Tsagkatakis, and P.~Tsakalides, ``1-bit tensor completion,''
	\emph{Electronic Imaging}, vol. 2018, no.~13, pp. 261--1, 2018.
	
	\bibitem{BLL2010}
	Y.~Baig, E.~M. Lai, and J.~Lewis, ``Quantization effects on compressed sensing
	video,'' in \emph{2010 17th International Conference on
		Telecommunications}.\hskip 1em plus 0.5em minus 0.4em\relax IEEE, 2010, pp.
	935--940.
	
	\bibitem{BLM11}
	L.~Baltrunas, M.~Kaminskas, B.~Ludwig, O.~Moling, F.~Ricci, A.~Aydin, K.-H.
	L{\"u}ke, and R.~Schwaiger, ``Incarmusic: Context-aware music recommendations
	in a car,'' in \emph{International Conference on Electronic Commerce and Web
		Technologies}.\hskip 1em plus 0.5em minus 0.4em\relax Springer, 2011, pp.
	89--100.
	
	\bibitem{Bhaskar16}
	S.~A. Bhaskar, ``Probabilistic low-rank matrix completion from quantized
	measurements,'' \emph{The Journal of Machine Learning Research}, vol.~17,
	no.~60, pp. 1--34, 2016.
	
	\bibitem{BNS16}
	S.~Bhojanapalli, B.~Neyshabur, and N.~Srebro, ``Global optimality of local
	search for low rank matrix recovery,'' in \emph{Advances in Neural
		Information Processing Systems}, 2016, pp. 3873--3881.
	
	\bibitem{BST14}
	J.~Bolte, S.~Sabach, and M.~Teboulle, ``Proximal alternating linearized
	minimization for nonconvex and nonsmooth problems,'' \emph{Mathematical
		Programming}, vol. 146, no. 1-2, pp. 459--494, 2014.
	
	\bibitem{BMR17}
	N.~I. Bruce, B.~Murthi, and R.~C. Rao, ``A dynamic model for digital
	advertising: The effects of creative format, message content, and targeting
	on engagement,'' \emph{Journal of Marketing Research}, vol.~54, no.~2, pp.
	202--218, 2017.
	
	\bibitem{CZ13}
	T.~Cai and W.-X. Zhou, ``A max-norm constrained minimization approach to 1-bit
	matrix completion,'' \emph{The Journal of Machine Learning Research},
	vol.~14, no.~1, pp. 3619--3647, 2013.
	
	\bibitem{CLKX13}
	S.~Chen, M.~R. Lyu, I.~King, and Z.~Xu, ``Exact and stable recovery of pairwise
	interaction tensors,'' in \emph{Advances in Neural Information Processing
		Systems}, 2013, pp. 1691--1699.
	
	\bibitem{CMH16}
	J.~Choi, J.~Mo, and R.~W. Heath, ``Near maximum-likelihood detector and channel
	estimator for uplink multiuser massive mimo systems with one-bit adcs,''
	\emph{IEEE Transactions on Communications}, vol.~64, no.~5, pp. 2005--2018,
	2016.
	
	\bibitem{CS2016}
	N.~Cohen and A.~Shashua, ``Convolutional rectifier networks as generalized
	tensor decompositions,'' in \emph{International Conference on Machine
		Learning}, 2016, pp. 955--963.
	
	\bibitem{DPBW14}
	M.~A. Davenport, Y.~Plan, E.~van~den Berg, and M.~Wootters, ``1-bit matrix
	completion,'' \emph{Information and Inference}, vol.~3, no.~3, pp. 189--223,
	2014.
	
	\bibitem{FL18}
	S.~Friedland and L.-H. Lim, ``Nuclear norm of higher-order tensors,''
	\emph{Mathematics of Computation}, vol.~87, no. 311, pp. 1255--1281, 2018.
	
	\bibitem{FGTL14}
	Y.~Fu, J.~Gao, D.~Tien, and Z.~Lin, ``Tensor {LRR} based subspace clustering,''
	in \emph{2014 International Joint Conference on Neural Networks
		(IJCNN)}.\hskip 1em plus 0.5em minus 0.4em\relax IEEE, 2014, pp. 1877--1884.
	
	\bibitem{GWWC18}
	P.~Gao, R.~Wang, M.~Wang, and J.~H. Chow, ``Low-rank matrix recovery from
	noisy, quantized and erroneous measurements,'' \emph{IEEE Transactions on
		Signal Processing}, vol.~66, no.~11, pp. 2918--2932, 2018.
	
	\bibitem{SBK01}
	A.~S. Georghiades, B.~Peter~N, and K.~David~J, ``From few to many: Illumination
	cone models for face recognition under variable lighting and pose,''
	\emph{IEEE Transactions on Pattern Analysis \& Machine Intelligence},
	vol.~23, no.~6, pp. 643--660, 2001.
	
	\bibitem{GPY19}
	N.~Ghadermarzy, Y.~Plan, and O.~Yilmaz, ``Learning tensors from partial binary
	measurements,'' \emph{IEEE Transactions on Signal Processing}, vol.~67,
	no.~1, pp. 29--40, 2019.
	
	\bibitem{Harshman70}
	R.~A. Harshman \emph{et~al.}, ``Foundations of the parafac procedure: Models
	and conditions for an" explanatory" multimodal factor analysis,'' 1970.
	
	\bibitem{KNSE19}
	S.~Khobahi, N.~Naimipour, M.~Soltanalian, and Y.~C. Eldar, ``Deep signal
	recovery with one-bit quantization,'' in \emph{ICASSP 2019-2019 IEEE
		International Conference on Acoustics, Speech and Signal Processing
		(ICASSP)}.\hskip 1em plus 0.5em minus 0.4em\relax IEEE, 2019, pp. 2987--2991.
	
	\bibitem{KB09}
	T.~G. Kolda and B.~W. Bader, ``Tensor decompositions and applications,''
	\emph{SIAM Review}, vol.~51, no.~3, pp. 455--500, 2009.
	
	\bibitem{LHK05}
	K.-C. Lee, J.~Ho, and D.~J. Kriegman, ``Acquiring linear subspaces for face
	recognition under variable lighting,'' \emph{IEEE Transactions on Pattern
		Analysis \& Machine Intelligence}, vol.~27, no.~5, pp. 684--698, 2005.
	
	\bibitem{LZLL19}
	B.~Li, X.~Zhang, X.~Li, and H.~Lu, ``Tensor completion from one-bit
	observations,'' \emph{IEEE Transactions on Image Processing}, vol.~28, no.~1,
	pp. 170--180, 2019.
	
	\bibitem{LZZZJ15}
	R.~Li, W.~Zhang, Y.~Zhao, Z.~Zhu, and S.~Ji, ``Sparsity learning formulations
	for mining time-varying data,'' \emph{IEEE Transactions on Knowledge and Data
		Engineering}, vol.~27, no.~5, pp. 1411--1423, 2015.
	
	\bibitem{LTSM17}
	Y.~Li, C.~Tao, G.~Seco-Granados, A.~Mezghani, A.~L. Swindlehurst, and L.~Liu,
	``Channel estimation and performance analysis of one-bit massive mimo
	systems,'' \emph{IEEE Transactions on Signal Processing}, vol.~65, no.~15,
	pp. 4075--4089, 2017.
	
	\bibitem{MTO18}
	K.~Maruhashi, M.~Todoriki, T.~Ohwa, K.~Goto, Y.~Hasegawa, H.~Inakoshi, and
	H.~Anai, ``Learning multi-way relations via tensor decomposition with neural
	networks,'' in \emph{Thirty-Second AAAI Conference on Artificial
		Intelligence}, 2018.
	
	\bibitem{PV2013}
	Y.~Plan and R.~Vershynin, ``Robust 1-bit compressed sensing and sparse logistic
	regression: A convex programming approach,'' \emph{IEEE Transactions on
		Information Theory}, vol.~59, no.~1, pp. 482--494, 2013.
	
	\bibitem{REC13}
	A.~Reinhardt, F.~Englert, and D.~Christin, ``Enhancing user privacy by
	preprocessing distributed smart meter data,'' in \emph{Proc. Sustainable
		Internet and ICT for Sustainability (SustainIT)}, 2013, pp. 1--7.
	
	\bibitem{RM2014}
	E.~Richard and A.~Montanari, ``A statistical model for tensor {PCA},'' in
	\emph{Advances in Neural Information Processing Systems}, 2014, pp.
	2897--2905.
	
	\bibitem{SK03}
	H.~S. Sahambi and K.~Khorasani, ``A neural-network appearance-based 3-d object
	recognition using independent component analysis,'' \emph{IEEE Transactions
		on Neural Networks}, vol.~14, no.~1, pp. 138--149, 2003.
	
	\bibitem{SL2015}
	M.~Slawski and P.~Li, ``b-bit marginal regression,'' in \emph{Advances in
		Neural Information Processing Systems}, 2015, pp. 2062--2070.
	
	\bibitem{TS14}
	R.~Tomioka and T.~Suzuki, ``Spectral norm of random tensors,'' \emph{arXiv
		preprint arXiv:1407.1870}, 2014.
	
	\bibitem{Tucker66}
	L.~R. Tucker, ``Some mathematical notes on three-mode factor analysis,''
	\emph{Psychometrika}, vol.~31, no.~3, pp. 279--311, 1966.
	
	\bibitem{WWX18}
	R.~Wang, M.~Wang, and J.~Xiong, ``Data recovery and subspace clustering from
	quantized and corrupted measurements,'' \emph{IEEE Journal of Selected Topics
		in Signal Processing}, vol.~12, no.~6, pp. 1547--1560, 2018.
	
	\bibitem{XY13}
	Y.~Xu and W.~Yin, ``A block coordinate descent method for regularized
	multiconvex optimization with applications to nonnegative tensor
	factorization and completion,'' \emph{SIAM Journal on Imaging Sciences},
	vol.~6, no.~3, pp. 1758--1789, 2013.
	
	\bibitem{XYH13}
	Y.~Xu, R.~Hao, W.~Yin, and Z.~Su, ``Parallel matrix factorization for low-rank
	tensor completion,'' \emph{arXiv preprint arXiv:1312.1254}, 2013.
	
	\bibitem{ZJWB09}
	G.~Zhang, J.~Jia, T.-T. Wong, and H.~Bao, ``Consistent depth maps recovery from
	a video sequence,'' \emph{IEEE Transactions on pattern analysis and machine
		intelligence}, vol.~31, no.~6, pp. 974--988, 2009.
	
	\bibitem{ZYJ14}
	L.~Zhang, J.~Yi, and R.~Jin, ``Efficient algorithms for robust one-bit
	compressive sensing,'' in \emph{International Conference on Machine
		Learning}, 2014, pp. 820--828.
	
	\bibitem{ZWZM13}
	X.~Zhang, D.~Wang, Z.~Zhou, and Y.~Ma, ``Simultaneous rectification and
	alignment via robust recovery of low-rank tensors,'' in \emph{Advances in
		Neural Information Processing Systems}, 2013, pp. 1637--1645.
	
	\bibitem{ZWL15}
	T.~Zhao, Z.~Wang, and H.~Liu, ``A nonconvex optimization framework for low rank
	matrix estimation,'' in \emph{Advances in Neural Information Processing
		Systems}, 2015, pp. 559--567.
	
	\bibitem{ZZW16}
	S.~Zhe, K.~Zhang, P.~Wang, K.-c. Lee, Z.~Xu, Y.~Qi, and Z.~Ghahramani,
	``Distributed flexible nonlinear tensor factorization,'' in \emph{Advances in
		Neural Information Processing Systems}, 2016, pp. 928--936.
	
\end{thebibliography}


\end{document}